\documentclass[a4paper,11pt]{article}
\usepackage{times}

\usepackage[english]{babel}
\usepackage[margin=1in]{geometry}
\usepackage{hyperref}
\usepackage{caption}
\usepackage{xcolor}
\usepackage{subcaption}
\usepackage[nolist]{acronym}
\usepackage{enumitem}
\usepackage{wrapfig}
\usepackage[capitalise]{cleveref}
\usepackage{textcomp}
\usepackage{graphicx}
\usepackage{url}
\usepackage{threeparttable}

\usepackage{pythonhighlight}
\usepackage{booktabs}

\providecommand{\Description}[1]{}

\crefname{figure}{Figure}{Figures}

\title{Leveraging Vision-Language Models for Visual Grounding and Analysis of Automotive UI }

\author{Benjamin Raphael Ernhofer$^*$$^+$, Daniil Prokhorov$^*$$^+$,  Jannica Langner$^*$ and Dominik Bollmann$^*$ \\ \\
\centerline{\small \em $^*$ SPARKS Solutions GmbH, Ingolstadt, Germany} \\
\centerline{\small \em $^+$ Equal contribution} \\ \\
}
\date{} 

\begin{document}

\maketitle

\pagestyle{myheadings}

\begin{abstract}
Modern automotive infotainment systems necessitate intelligent and adaptive solutions to manage frequent User Interface (UI) updates and diverse design variations.
This work introduces a  vision-language framework to facilitate the understanding of and interaction with automotive UIs, enabling seamless adaptation across different UI designs.
To support research in this field, AutomotiveUI-Bench-4K, an open-source dataset comprising 998 images with 4,208 annotations, is also released. 
Additionally, a data pipeline for generating training data is presented.

A Molmo-7B-based model is fine-tuned using Low-Rank Adaptation (LoRa), incorporating generated reasoning along with visual grounding and evaluation capabilities.
The fine-tuned Evaluative Large Action Model (ELAM) achieves strong performance on AutomotiveUI-Bench-4K (\href{https://huggingface.co/sparks-solutions/ELAM-7B}{\underbar{model}} and \href{https://huggingface.co/datasets/sparks-solutions/AutomotiveUI-Bench-4K}{\underbar{dataset}} are available on Hugging Face).
The approach demonstrates strong cross-domain generalization, including a +5.6\% improvement on ScreenSpot over the baseline model.
An average accuracy of 80.8\% is achieved on ScreenSpot, closely matching or surpassing specialized models for desktop, mobile, and web, despite being trained primarily on the automotive domain.
This research investigates how data collection and subsequent fine-tuning can lead to AI-driven advancements in automotive UI understanding and interaction.
The applied method is cost-efficient, and fine-tuned models can be deployed on consumer-grade GPUs.
 
\vspace{0.5cm}

\end{abstract}
\hrulefill
\begin{acronym}
    \acro{adas}[ADAS]{Advanced Driver-Assistance Systems}
    \acro{ecu}[ECU]{Electronic Control Unit}
    \acro{elam}[ELAM]{Evaluative Large Action Model}
    \acro{gui}[GUI]{Graphical User Interface}
    \acro{hil}[HiL]{Hardware-in-the-Loop}
    \acro{hmi}[HMI]{Human-Machine Interface}
    \acro{lam}[LAM]{Large Action Model}
    \acro{llm}[LLM]{Large Language Model}
    \acro{lora}[LoRa]{Low-Rank Adaptation}
    \acro{oem}[OEM]{Original Equipment Manufacturer}
    \acro{som}[SoM]{Set-of-Mark}
    \acro{ui}[UI]{User Interface}
    \acro{vlm}[VLM]{Visual Language Model}
    \acro{ocr}[OCR]{Optical Character Recognition}
\end{acronym}
\section{Introduction}

Automotive infotainment systems are rapidly evolving, characterized by increasing complexity, dynamic interfaces, and personalization \cite{meixner2017automotive}. With manufacturers frequently deploying over-the-air updates and diverse \ac{ui} designs becoming prevalent across models, these systems demand intelligent and adaptive solutions capable of handling significant variations \cite{yin2018automated}. This constant evolution necessitates intelligent systems capable of dynamically interpreting the visual layout and semantic meaning of interfaces, moving beyond reliance on fixed structural assumptions for \ac{ui} validation.

\acp{vlm} offer a promising approach by integrating computer vision with natural language understanding, enabling systems to interpret visual information and user intent in a more human-like manner \cite{li2025benchmark}. While these models have demonstrated success in understanding and interacting with interfaces in domains such as desktop, mobile, and web \cite{cheng2024seeclick,gou2025uground}, their application to the distinct environment of automotive infotainment systems remains significantly underexplored. This represents a research gap, given the unique demands and rapid evolution of in-vehicle interfaces. Applying \acp{vlm} effectively within the automotive context introduces non-trivial challenges. Automotive \acp{ui} exhibit vast heterogeneity across car models and manufacturers, featuring custom icons, menus, and interaction paradigms \cite{meixner2017automotive}. Therefore, a robust model must generalize effectively across disparate screen layouts and styles while maintaining high precision in understanding on-screen elements and their context. Moreover, sophisticated interaction often requires reasoning about the \ac{ui} state, extending beyond simple object detection or element localization.

To bridge this gap, a highly adaptable vision-language framework is introduced, designed for understanding, interacting, and validating automotive infotainment systems and enabling seamless adaptation across different \ac{ui} designs. The key contributions of this work include:
\begin{enumerate}
\item The fine-tuned Molmo-based \acl{elam} \textbf{\acs{elam}-7B}, optimized for automotive \ac{ui} understanding, capable of processing visual input and natural language test action and evaluation instructions.
\item The release of \textbf{AutomotiveUI-Bench-4K}, an open-source dataset featuring 998 infotainment images with 4,208 annotations, hosted on Hugging Face, to serve as a benchmark and foster research in this domain.
\item A \textbf{synthetic data generation pipeline} developed to enhance model performance and generalization for small \acp{vlm} (7B or less) through parameter-efficient fine-tuning.
\end{enumerate}
Through empirical evaluation, the efficacy of the proposed framework is demonstrated. The fine-tuned model establishes a new performance benchmark on the AutomotiveUI-Bench-4K dataset and exhibits remarkable adaptability, significantly improving results on the cross-domain ScreenSpot task (+5.6\%). 
Its overall accuracy (80.8\% average) competes favorably with, and in some cases surpasses, models specifically designed for non-automotive domains (e.g., ShowUI \cite{lin2024showui}), as shown in \cref{tab:screenspot_results}.
The robustness of the model is highlighted by its performance even with a restricted training dataset, both in terms of size and domain.
It is important to note, however, that the automotive infotainment domain itself encompasses a diverse range of systems and functionalities.
The following sections cover the system architecture, the curation of the novel dataset, and the comprehensive experimental validation, collectively showcasing the significant potential of AI-driven methods for advancing automotive \ac{ui} understanding and interaction.
\section{Related Work}

This section reviews the literature relevant to automated \ac{ui} interaction and validation. It begins by examining the current landscape of \acp{vlm} and the emerging field of \acp{lam} applied to general \ac{ui} understanding. Subsequently, existing domain-specific datasets and common model adaptation techniques are analyzed, highlighting their limitations for automotive applications. The section concludes by discussing traditional validation methodologies to contextualize and define the specific research gaps addressed in this paper.

\subsection{Vision Language Models and UI Understanding}
\acp{vlm} have emerged as a significant area of research, bridging the gap between visual and textual information processing. These models integrate visual capabilities with \acp{llm} to handle complex tasks such as image captioning, visual question answering, and multimodal dialogue systems \cite{ghosh2024exploring}. Key contributions in this field include models like CLIP \cite{radford2021learning}, which demonstrates strong performance in zero-shot image classification by aligning visual and textual embeddings, and BLIP-2 \cite{li2023blip}, which introduces an efficient pre-training strategy using frozen image encoders and \acp{llm}. Other notable models include Flamingo \cite{alayrac2022flamingo}, which exhibits significant few-shot learning capabilities, and LLaVA \cite{liu2023visualinstructiontuning}, which enhances \acp{llm} for multimodal understanding through visual instruction tuning.

A significant advancement in this domain is the Molmo series \cite{deitke2024molmo}, which introduces approaches to multimodal understanding with notable capabilities in spatial reasoning and accurate localization. Unlike traditional \acp{vlm} that primarily rely on discrete vision encoders, Molmo models demonstrate enhanced capability in understanding spatial relationships and providing localization through their pointing mechanism. This pointing capability allows models to not only identify \ac{ui} elements but also provide exact coordinate information for their locations, representing a significant development for \ac{ui} interaction tasks. The Molmo series is trained on the comprehensive PixMo dataset \cite{deitke2024molmo}, which contains millions of image-text pairs with spatial annotations, including pointing data that enables models to ground textual descriptions to specific normalized pixel coordinates in images.

As \acp{vlm} evolved, models specializing in \ac{ui} interaction were developed, leading to the establishment of the term \acf{lam} \cite{wang2024large}. Comprehensive surveys \cite{zhang2024large,nguyen2024gui,wang2024gui} have addressed critical research questions concerning the development of \acp{lam} for \ac{ui} tasks. It has been highlighted that a specialized \ac{vlm} capable of accurate visual grounding is crucial for strong performance. Frameworks like ShowUI \cite{lin2024showui} and SeeClick \cite{cheng2024seeclick} leverage \acp{vlm} to map natural language instructions to \ac{ui} element interactions. Similarly, \acp{lam} such as OS-Atlas \cite{wu2024atlas} extend \acp{vlm} by incorporating action generation capabilities suitable for agentic \ac{ui} navigation tasks. Furthermore, smaller fine-tuned models like TinyClick \cite{pawlowski2024tinyclick} (based on Microsoft's Florence-2-Base \cite{xiao2023florence}) and UGround \cite{gou2025uground} (based on Qwen2-VL \cite{wang2024qwen2}) have been developed specifically for \ac{ui} tasks, often processing \ac{ui} images and interpreting actions to generate corresponding point coordinates.

The Molmo series differentiates itself from other \ac{ui}-focused models through its training methodology that incorporates both general multimodal understanding and specialized pointing capabilities. While most \acp{lam} are trained primarily on \ac{ui}-specific datasets, Molmo models benefit from the diverse PixMo dataset, which includes a smaller but significant portion of \ac{ui}-related data from the AndroidControl \cite{li2024effects} dataset alongside general visual content. This approach enables better generalization capabilities while maintaining strong performance on \ac{ui} tasks. The pointing mechanism in Molmo models provides more accurate spatial understanding compared to traditional region-based approaches, such as those relying on bounding boxes, used in other models.

However, these models are typically trained on general-purpose interfaces from desktop, mobile, and web domains. This limits their applicability to automotive \acp{ui}, which exhibit domain-specific design patterns and custom iconography. Notably, they lack explicit mechanisms for evaluating \ac{ui} states (e.g., verifying if a seat belt warning is displayed), an essential requirement for automotive validation. While the Molmo series demonstrates enhanced spatial reasoning capabilities through its pointing mechanism, it still faces the same domain adaptation challenges when applied to automotive interfaces with their unique visual characteristics and functional requirements.

\subsection{Domain-Specific UI Datasets and Model Adaptation}
Existing \ac{ui} datasets, including AMEX \cite{chai2024amex} and Android In the Wild \cite{rawles2023androidinthewild}, focus on mobile or web environments. These datasets primarily annotate \ac{ui} elements with interaction labels (e.g., ``tap the settings icon'') but rarely include evaluative statements. The PixMo dataset \cite{deitke2024molmo}, while comprehensive in its multimodal coverage with millions of image-text pairs and innovative pointing annotations, also follows this pattern. Its \ac{ui}-related subset focuses primarily on interaction rather than validation tasks. The dataset's strength lies in its rich spatial annotations that enable accurate localization, but it lacks the evaluation-centric labels necessary for automotive \ac{ui} validation scenarios.

In contrast, the proposed \textbf{AutomotiveUI-Bench-4K} introduces a dual-label structure (\textit{Test Action} and \textit{Expected Result}) to simulate real-world validation scenarios, aligning with the need for compliance checks. This approach differs fundamentally from existing datasets, including PixMo. While these datasets excel in interaction modeling, they do not address the critical evaluation dimension required for automotive \ac{ui} testing.

Parameter-efficient fine-tuning techniques such as \ac{lora} \cite{hu2022lora} have been used to adapt \acp{vlm} for \ac{ui} tasks, as seen in UGround \cite{gou2025uground}. However, these approaches focus solely on interaction and neglect evaluation. Likewise, prior synthetic data pipelines \cite{gou2025uground,pawlowski2024tinyclick} generate interaction-focused annotations. These adaptation methods are extended in this work by the introduction of a novel pipeline. This pipeline balances \textit{Test Actions} with \textit{Expected Results} to train evaluation-aware models.

\subsection{Traditional Automotive UI Validation and Research Gaps}
The validation of automotive \acp{ui} traditionally relies on methodologies like specification-based testing \cite{khan2012comparative} and \ac{hil} testing \cite{king2004hardware,kumar2020automotive}. Within \ac{hil} setups, visual assessment often employs \ac{ocr} and template matching \cite{2017FunctionalGT,vector_canoe_simulation}. These conventional approaches are fragile, struggling with visual variations from updates or themes. Furthermore, they lack a deep semantic understanding of interface elements \cite{smith2007overview,fu2024understanding}. Consequently, the maintenance overhead necessitates more intelligent solutions.
This work bridges two critical gaps in prior research:
\begin{enumerate}
	\item \textbf{Automotive-Specific VLMs}: Existing \acp{vlm}, including advanced models like the Molmo series with their spatial reasoning capabilities, are trained on generic \acp{ui}. This limits their generalization to automotive interfaces with custom iconography and layouts.
	\item \textbf{Evaluation-Centric Training}: Unlike most \acp{lam} that emphasize interaction (including Molmo's pointing mechanism for interaction tasks), an evaluation-focused training approach is introduced in this research, enabling analysis and validation of automotive \acp{ui}.
\end{enumerate}
Therefore, \acp{vlm} are leveraged to overcome the limitations of traditional methods, establishing a more robust and semantically aware framework for automotive \ac{ui} validation. The development and evaluation of this framework constitute the core of this work. Specific recommendations for its integration into production-ready \ac{hil} testing environments are not provided.
\section{Method}

The methodology for adapting \acp{vlm} for automated automotive \ac{ui} validation is outlined in this section. Domain adaptation challenges are addressed through an approach encompassing synthetic data generation and fine-tuning of baseline models, leading to the development of a specialized \ac{vlm} for accurate visual grounding and evaluation of automotive \acp{ui}.

\subsection{Domain Adaptation Challenges and Model Selection}

Extensive research exists on \acp{lam} and \ac{ui} agents. However, these models primarily focus on interactions and implicitly derive evaluative functions by inferring necessary actions. For the targeted application of validating requirements via natural language test cases, explicit evaluation of expected outcomes is necessary. Existing smaller fine-tuned models (7B or less), such as TinyClick \cite{pawlowski2024tinyclick} and UGround \cite{gou2025uground}, gather \ac{ui} data from mobile, desktop, and web applications. This limits their applicability to automotive \acp{ui}. Domain-specific adaptation is required due to the distinct design patterns, custom iconography, and varied display hardware properties of modern automotive systems.

Specifically, the following challenges necessitated the proposed methodology:

\begin{enumerate}
    \item \textbf{Adapt \acp{vlm} for the domain of automotive \ac{ui}}: Modern automotive systems exhibit diverse \ac{ui} designs influenced by brand, platform, driving mode, and display hardware properties. Icons pose a particular challenge, as little overlap exists with generic \acp{ui} from desktop, mobile, or web applications.
    
    \item \textbf{Assert evaluation capability}: Explicit evaluation is not considered in prior work. To evaluate requirements or specifications, expected results must be prompted. These results require visual grounding and subsequent evaluation.
    
    \item \textbf{Ensure onsite deployment and control of models}:
    \begin{itemize}
        \item Strict information security requirements are often held by \acp{oem}. The leveraging of proprietary models such as OpenAI GPT-4o or Google Gemini Pro 2.0 presents challenges due to data privacy concerns and potential leakage of sensitive information to external servers.
        \item Fine-tuning such models presents another challenge. While OpenAI and Google provide fine-tuning services for training and deployment, the details of the training process and the fine-tuned model itself remain closed source.
        \item Challenging validation and prompt adaptation iterations to ensure intended functionality arise from the discontinuation of closed-source models.
        \item The operation of models on local servers or even consumer-grade GPUs is highly desirable and necessary for a cost-effective solution.
    \end{itemize}
\end{enumerate}
To address these challenges, \textbf{\ac{elam}} is introduced, extending existing \acp{lam} through the integration of explicit evaluation capabilities. The model is trained with synthetic data generated by an accompanying pipeline (\cref{sec:synthetic_data_pipeline}). This pipeline is designed to address the unique characteristics of automotive \acp{ui} and the need for evaluative statements.

To establish a suitable baseline model for fine-tuning, four existing \acp{vlm} were evaluated: \textit{TinyClick} (0.27B) \cite{pawlowski2024tinyclick}, \textit{Molmo-7B-D-0924} \cite{deitke2024molmo}, \textit{UGround-V1-7B} (Qwen2-VL-based) \cite{gou2025uground}, and \textit{InternVL2\_5-8B} \cite{chen2025internvl25}. TinyClick and UGround are specifically designed for \ac{ui} interaction and point coordinate generation. However, their broader capabilities and suitability as a robust foundation for evaluation were limited. General-purpose \acp{vlm}, Molmo and InternVL2.5, have also been trained on \ac{ui} data. The best performance (\cref{tab:hmi_bench_results}) among the considered models was achieved by \textit{Molmo-7B-D-0924} when evaluated on the proposed AutomotiveUI-Bench-4k dataset. Consequently, it was chosen as the foundation of \ac{elam}-7B.

\subsection{Synthetic Data Pipeline} \label{sec:synthetic_data_pipeline}

Object-level captioning for automotive \acp{ui} presents significant challenges due to the need to incorporate functional, positional, and visual attributes to uniquely identify each element. Differentiating between \textit{Test Actions} and \textit{Expected Results} is required. Both necessitate object-level captions describing either the command initiating element actuation or a statement to be evaluated for the current \ac{ui}. Furthermore, for \textit{Expected Results}, a binary status (passed or failed) must be assigned, reflecting the validity of the expectation within the image. Empirical observation revealed that human annotators tend to under-represent failed expectations, likely due to the cognitive challenge of formulating incorrect expectations for a displayed \ac{ui} image. It is also important to note that having human experts create data annotations is a time-consuming and expensive process. Therefore, a hybrid approach, leveraging human expertise for \ac{ui} element selection via bounding box annotations combined with automated caption generation by large teacher models (e.g., GPT-4o, Gemini 2.0 Flash Thinking), offers a practical balance between annotation quality and time investment. Automated \ac{ui} element extraction methods, such as OmniParser V1.5 \cite{lu2024omniparserpurevisionbased} and a custom YOLOv8 \cite{Jocher_Ultralytics_YOLO_2023} fine-tuned to automotive \ac{ui}, were found to be insufficiently reliable, frequently resulting in misaligned bounding box detections that negatively impacted the correctness of generated captions.

\begin{figure*}[ht]
    \centering
    \includegraphics[width=\linewidth]{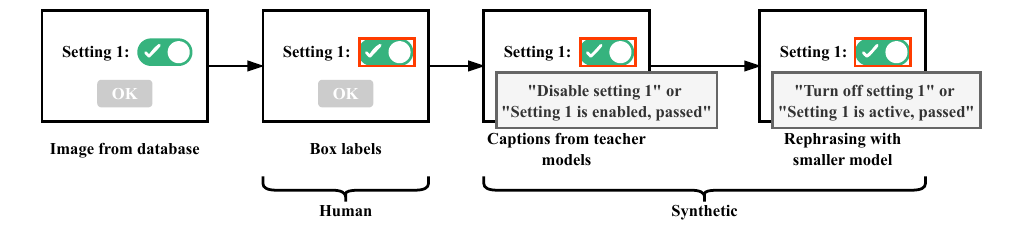}
    \caption{Synthetic data generation pipeline}
    \Description{The image is a diagram titled ``Synthetic data generation pipeline'' showing a four-step process from left to right. Each step is represented by a rectangular box connected by arrows. The first two boxes are grouped under ``Human'', and the last two under ``Synthetic''.
        Step 1: ``Image from database'': A box shows a toggle switch labeled ``Setting 1:'' in the `on' position and a button labeled ``OK''.
        Step 2: ``Box labels'': Similar to Step 1, with the toggle switch outlined in a red box.
        Step 3: ``Captions from teacher models'': Again with the toggle switch outlined, this box also contains text: ''Disable setting 1'' or ``Setting 1 is enabled, passed''.
        Step 4: ``Rephrasing with smaller model'': Similar toggle switch outlined, and text: ``Turn off setting 1'' or ``Setting 1 is active, passed''.
        The diagram illustrates a synthetic data generation pipeline starting with a database image, then human box labeling, followed by synthetic caption generation and rephrasing.
    }
    \label{fig:synthetic_data_generation_pipeline}
\end{figure*}

\begin{figure*}[ht]
    \centering
    \begin{subfigure}{0.48\textwidth}
        \centering
        \includegraphics[width=\linewidth]{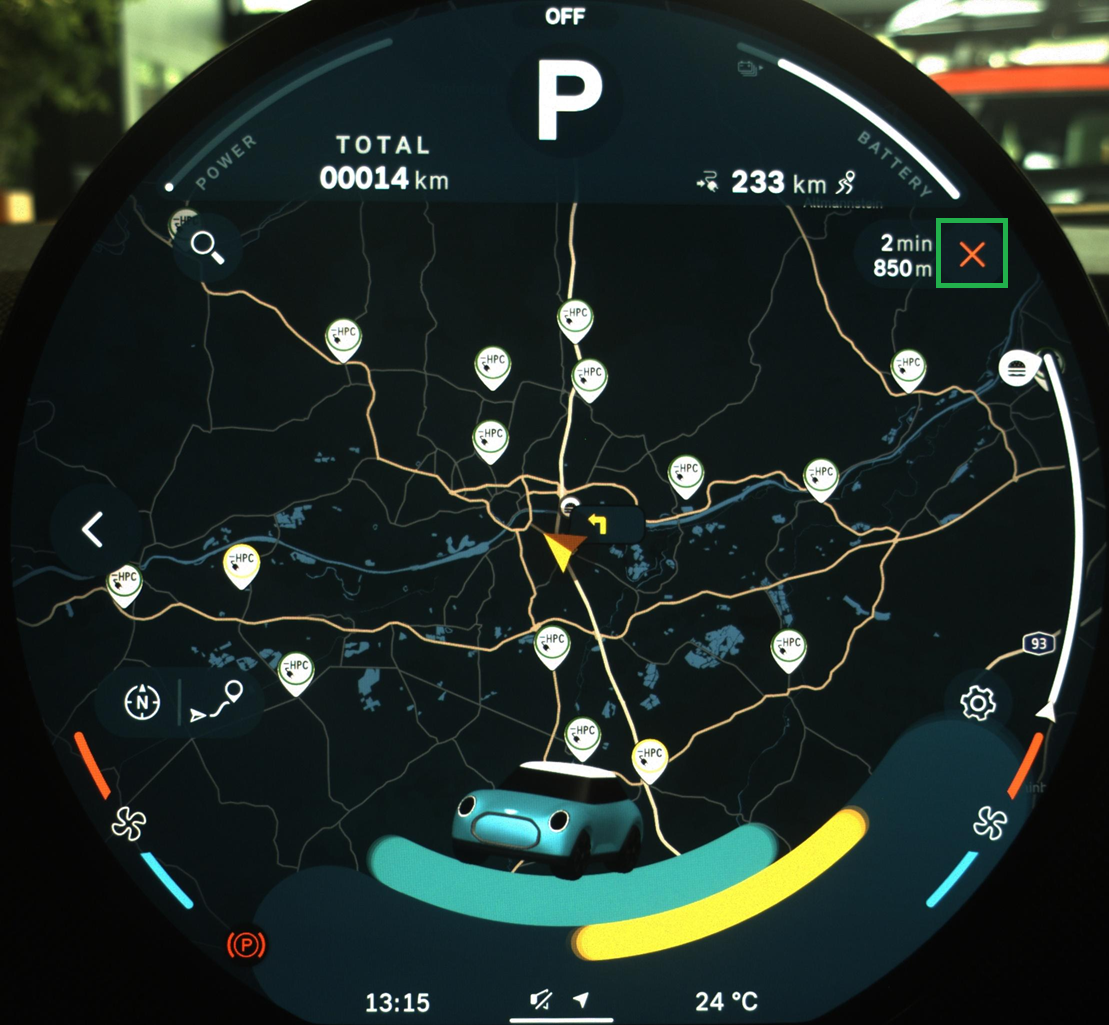} 
        \caption{Synthetic \textit{Test Action} \textbf{}}
        \label{fig:train_sample_mini_test_action}
        \Description{Image of a navigation route in a Mini Cooper infotainment system. The top right corner shows a widget that contains information about the estimated time and distance left for the current route. Next to the two metrics is a red cross for closing the route guidance. This UI element is highlighted by a red bounding box.}
    \end{subfigure}\hfill
    \begin{subfigure}{0.48\textwidth}
        \centering
        \includegraphics[width=\linewidth]{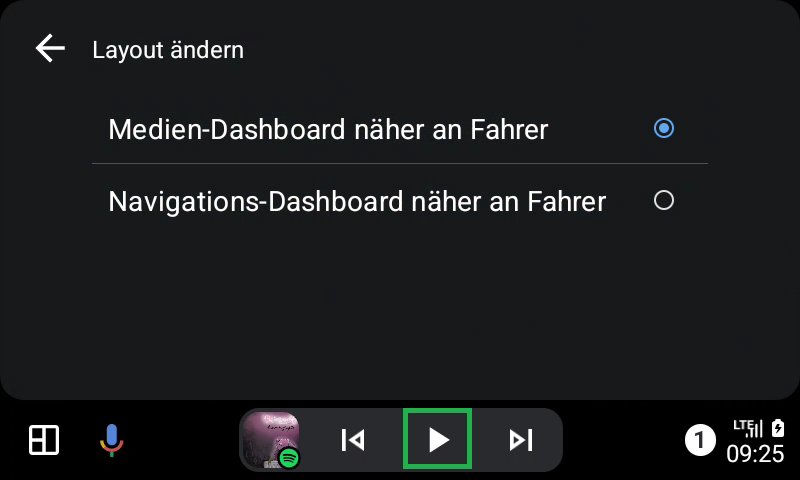} 
        \caption{Synthetic \textit{Expected Result}}
        \label{fig:train_sample_android_failed_exp_res}
        \Description{Image of an Android Auto system in which the media control shows a play icon. The play icon is highlighted using a green bounding box.}
    \end{subfigure}
    \caption{Examples of synthetic data for training, illustrating (a) a test action \textbf{``Tap the 'X' button to cancel the displayed route guidance information on the map.''} with prior reasoning \textit{``The requested element is an ``X'' icon located in the top right corner of the map display, semantically representing a close or cancel action, positioned next to the text ``2 min'' and ``850 m,'' which likely refers to route guidance information, with the parent element being the map view.''} and (b) a failed expected result \textbf{``The media control button in the bottom bar is expected to be a pause icon, indicating that media is currently playing. $\,\to\,$ Failed''} with prior reasoning \textit{``The requested element is located in the bottom media control bar and is identified as a play symbol, which semantically represents the action to start or resume media playback. The evaluation highlights that there could be a misunderstanding regarding the icon's function, as it may be incorrectly expected to represent a pause action instead. This confusion emphasizes the importance of clear iconography in user interface design to ensure users can easily understand the intended actions associated with each control.''} in different infotainment systems.}
    \label{fig:two_training_samples}
\end{figure*}

The synthetic data pipeline (\cref{fig:synthetic_data_generation_pipeline}) was used to train \ac{elam}-7B. Bounding box regions from the dataset were prompted using the \ac{som} technique \cite{yang2023setofmark}. Inspired by UGround \cite{gou2025uground}, a bounding box indicated by an arrow was utilized. For \textit{Test Actions}, prompts were designed as short yet comprehensive instructions that unambiguously map to a single \ac{ui} element (\cref{fig:train_sample_mini_test_action}). Prior to generating each \textit{Test Action}, the model was instructed to provide a detailed reasoning of the process for identifying the target \ac{ui} element, including its semantic, positional, and visual characteristics, as well as any relevant relationships to other elements, such as the current context or list headings (\cref{fig:two_training_samples}). Utilized prompts can be found in \cref{sec:synthetic_data_gen_prompts_gemini}.

To improve the generation of \textit{Expected Results}, the process was modified to include a prior successful test action leading to the current menu or \ac{ui} state (\cref{fig:train_sample_android_failed_exp_res}). Specifically, the model first generated a prior test action, followed by reasoning for the expectation and its ``passed/failed'' conclusion. Initially, this approach exhibited the same bias as human annotators and primarily generated passed validations. This issue was mitigated by explicitly prompting for evaluations that entail a failed result (\cref{promt:synt_expected_result_failed}). After manually reviewing a small set of samples for GPT-4o, Gemini 2.0 Flash Thinking and Claude 3.5 Sonnet, Gemini (\textit{gemini-2.0-flash-thinking-exp-01-21}) was selected to generate the training data. It was chosen for its better adherence to the required output structure, which was necessary for automatic parsing.

To mitigate potential overfitting on the specific linguistic style of the teacher model, the text for reasoning, test actions, and expectations were rephrased using a smaller model (\textit{gpt-4o-mini-2024-07-18}). This rephrasing step is crucial for ensuring robustness and generalizability when processing inputs from real-world test case data. The teacher model, configured with a temperature of zero to minimize hallucinations, exhibited a tendency towards repetitive use of verbs and sentence structures, necessitating this diversification strategy. The contribution of rephrasing is explored in \cref{tab:ablation_reasoning_rephrasing}.

\subsection{Model Fine-Tuning}

For efficient training of baseline models, \ac{lora} \cite{hu2022lora} was utilized. This parameter-efficient fine-tuning method injects small, trainable matrices into the model's layers, allowing for significant improvements without retraining the entire large model. This approach offers benefits such as accelerated training, reduced training data needs, a smaller storage footprint for the adapters, lower memory requirements, and the potential for dynamic adapter swapping in deployment, which eliminates the need to load different large models for various tasks.

As previously established, \textit{Molmo-7B-D-0924} was chosen as the foundation for \ac{elam} due to its strong performance on the AutomotiveUI-Bench-4k dataset among evaluated baseline models. This foundation provided a robust starting point. Fine-tuning with LoRA then enabled efficient adaptation to the specific tasks of automotive UI validation, leveraging a synthetically generated dataset that explicitly includes both \textit{Test Actions} and \textit{Expected Results} with their corresponding passed/failed states. This fine-tuning specifically aims to imbue \ac{elam}-7B with the evaluation capabilities lacking in prior interaction-focused \acp{lam}, and to shift the learned domain to automotive \ac{ui}. Further details regarding this fine-tuning can be found in \cref{sec:experimental_config}.

\section{Open Source Infotainment Validation Dataset: AutomotiveUI-Bench-4K}
\label{sec:novel_automotive_ui_validation_dataset}

Despite the availability of numerous datasets for benchmarking \acp{lam} in desktop, mobile, and web \acp{ui} \cite{rawles2023androidinthewild,wu2024atlas,wu2023webui,website-screenshots_dataset,zheng2024agentstudio,chai2024amex,gao2024mobileviews}, a significant domain gap remains in their application to automotive \acp{ui}. To address this limitation, \textbf{AutomotiveUI-Bench-4K} is introduced and released as a new dataset specifically designed for this domain.
This dataset comprises 998 images with 4,208 annotations from 15 brands, including Audi, BMW, Ford, Porsche, and Tesla.
This diverse collection of modern vehicle \acp{ui} was annotated by experienced professional software test engineers specializing in \ac{hmi} evaluation. 
In addition to images from automotive \acp{oem}, the dataset also incorporates screenshots of Apple CarPlay and Google Android Auto.
Images from \acp{oem} were captured in 4K resolution using cameras and sourced from either a research vehicle fleet or through collaboration with local car dealerships.
To ensure consistent image perspective and focus on the relevant \ac{ui} elements, the \ac{oem} images were rectified and cropped to the active display area.
Android Auto and CarPlay screenshots were directly exported.
The systems featured in this dataset are exclusively touchscreen-based.
While some systems include auxiliary physical buttons, these were not considered in the analysis due to their primarily static functionality.
 This focused approach is justified by the increasing ubiquity of touchscreen interfaces as the dominant \ac{hmi} in modern vehicles. By concentrating solely on this technology, the dataset provides a unified corpus for analysis. The resulting data homogeneity is ideal for robust comparative analysis and the development of  \acp{vlm} specifically for this contemporary \ac{ui} paradigm.

\begin{figure*}[ht]
    \centering
    \includegraphics[width=0.95\linewidth]{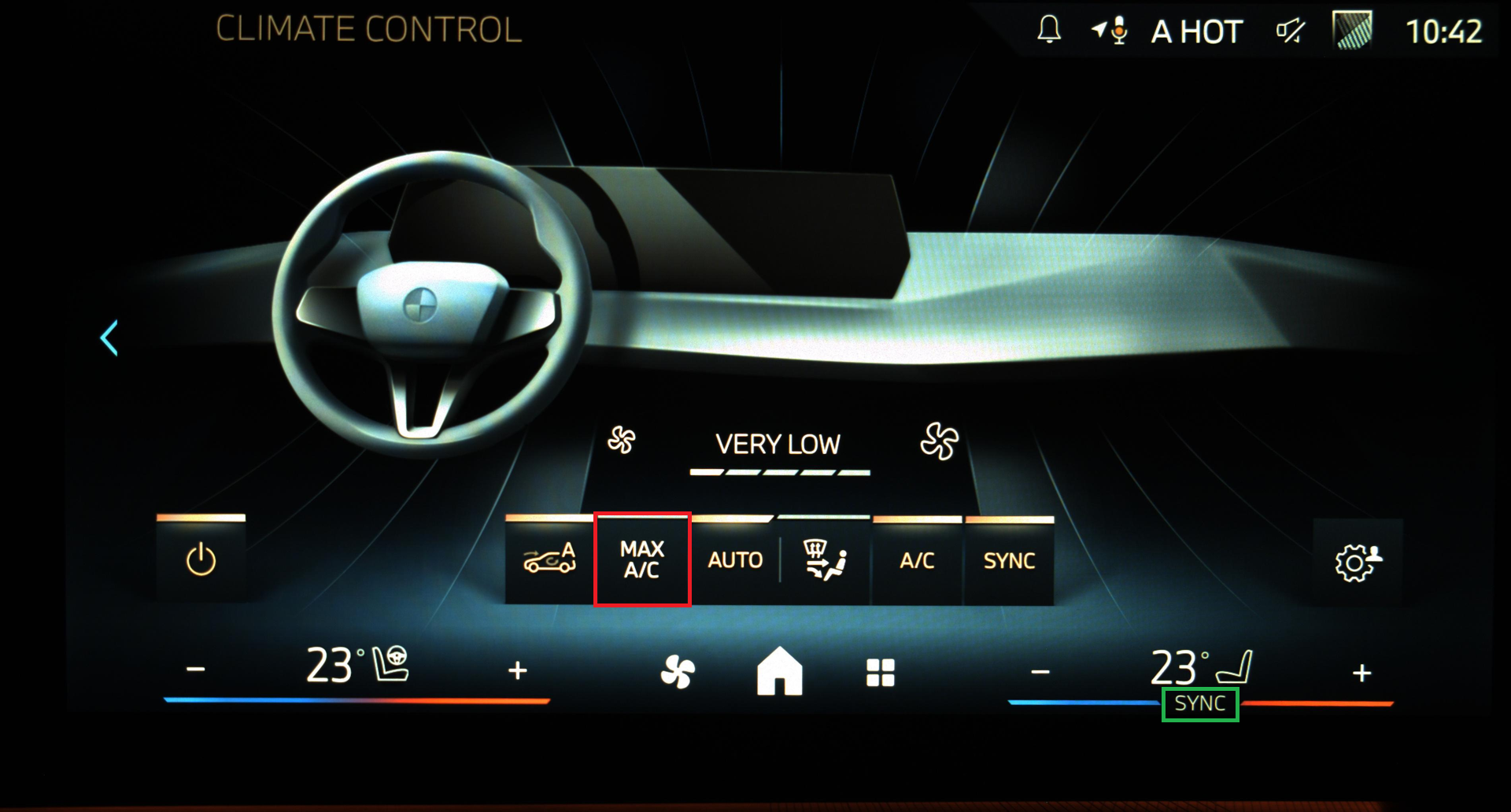} 
    \caption{Example for a \textit{Test Action} (red) \textit{``set A/C to max''} and for an \textit{Expected Result }(green) \textit{``Passenger's climate zone is synced to driver'' (Passed)} }
    \Description{This image shows the climate control screen of a BMW iX2 in English with a dark background and light text. The first highlighted element is a rectangular button outlined in red, labeled "MAX A/C", located in the center of the screen among other climate control buttons. Below and to the right of these buttons within the control panel for the passenger's side, the second highlighted element is the text label ``SYNC'' outlined in green.}
    \label{fig:eval_bmw_example}
\end{figure*}

Although the dataset primarily comprises systems from recent model years, the system representing the earliest model year is a 2018 Volkswagen Tiguan.
Languages that were included are German and English. Instructions and evaluation requests are written in English.
German text was either translated or directly quoted.
Real-world test engineering documentation often exhibits a pattern of short sentences and occasional grammatical deviations.
Therefore, to create a dataset that realistically represents this domain, it is crucial to preserve these stylistic features in the preprocessing steps.
\Cref{tab:eval_dataset_stats} provides further information about the data distribution. 
The actions and results are unrelated, as test actions can lead to different \ac{ui} states, and screen trajectories are not considered.
There are two annotation classes in the dataset:

\begin{enumerate}[label=\arabic*.]
	\item \textbf{Test Action:} Describes an interaction with a single \ac{ui} element as an imperative sentence ("set A/C to max" in \cref{fig:eval_bmw_example}).
	\item \textbf{Expected Result:} Represents a testable expectation, defined per image, focusing on the required appearance, context, or state  of \ac{ui} elements within that image. This expectation is evaluated to determine a passed/failed outcome. ("Passenger's climate zone is synced to driver" in \cref{fig:eval_bmw_example})
\end{enumerate}

\begin{table*}[ht]
	\centering
	\caption{Label distribution in AutomotiveUI-Bench-4K}
	\label{tab:eval_dataset_stats}
	\begin{tabular}{llccc}
		\toprule
		Category & Subcategory & Total & EN & DE \\
		\midrule
		Images & - & 998 & 454 & 544 \\
		\hline
		Annotations & - & 4,208 & 1,988 & 2,220 \\
		\hline
		Test Action & -  & 2,269 & 1,059 & 1,210 \\
		\hline
		Expected Result & Total & 1,939 & 929 & 1,010 \\
		Expected Result & Passed & 1,375 & 662 & 713 \\
		Expected Result & Failed & 564 & 267 & 297 \\
		\bottomrule
	\end{tabular} 
\end{table*}

\section{Experiments}

This section details the experimental procedure for fine-tuning and evaluating \ac{elam}. The following subsections cover the specific configuration used for training, the performance of the fine-tuned model against several baselines on the AutomotiveUI-Bench-4K and ScreenSpot datasets. The impact of different training data preprocessing strategies is explored through ablation studies, the relationship between language semantics and model performance is investigated using t-SNE, and a detailed visual error analysis is conducted.

\subsection{Experimental Configuration}\label{sec:experimental_config}

To train \ac{elam}, \ac{lora} was applied to all linear layers within the Molmo architecture.
Training was conducted for 2 epochs with a global batch size of 128, achieved using a local batch size of 8 and gradient accumulation.
A learning rate of $1\mathrm{e}{-4}$ was employed.
An optimal \ac{lora} rank $r=64$ was selected based on an ablation study (\cref{tab:ablation_lora}), with the $\alpha$ parameter set to $r$ and a fixed dropout rate of $0.05$. ModelScope's SWIFT framework \cite{zhao2024swiftascalablelightweightinfrastructure} was used to conduct fine-tuning.

Training data for \ac{elam} was generated using the data pipeline described in \cref{sec:synthetic_data_pipeline}.
Gemini (\textit{gemini-2.0-flash-thinking-exp-01-21}) was selected for training dataset generation due to its superior adherence to structured output compared to other models.
The detailed prompts used for generating the training datasets are displayed in \cref{promt:synt_test_action,promt:synt_expected_result_passed,promt:synt_expected_result_failed}.

Generated text for reasoning, test actions, and expectations was rephrased using a smaller model (\textit{gpt-4o-mini-2024-07-18}).
The synthetic dataset comprises 17,708 annotations across 6,230 images.
This includes 5,952 \textit{Test Actions}, 6,190 passed and 5,566 failed \textit{Expected Results}.
Prompt templates utilized for training, evaluation, and inference are detailed in \cref{sec:elam-prompt-templates}.
Model training was performed on two NVIDIA H100 80 GB PCIe GPUs.
For evaluation experiments on the AutomotiveUI-Bench-4K dataset, an NVIDIA GeForce RTX 4090 GPU with 24 GB VRAM was used.

Performance of the models was assessed using accuracy metrics, all expressed as percentages. For the visual grounding task, two distinct accuracy metrics were employed: $TA_{vg}$ represents the visual grounding accuracy for \textit{Test Actions}, while $ER_{vg}$ denotes the visual grounding accuracy for \textit{Expected Results}. For both $TA_{vg}$ and $ER_{vg}$, accuracy was determined by verifying that the generated point or the centroid of the predicted box was contained within the annotated ground truth box.
Furthermore, for the evaluation of \textit{Expected Results}, $ER_{evl}$ was used, representing the classification accuracy (passed or failed).
To facilitate a fine-grained analysis of language-specific performance, the image dataset was categorized into English and German subsets. Corresponding language-specific metrics are indicated by the superscripts $^{EN}$ and  $^{DE}$ for English and German UIs, respectively.
In cases where a generated evaluation could not be parsed from the model's response, it was assigned the inverse of the ground truth value.

\subsection{Evaluating Performance with AutomotiveUI-Bench-4K}

\begin{figure*}[ht]
	\centering
	\begin{subfigure}{0.44\textwidth}
		\centering
		\includegraphics[width=\linewidth]{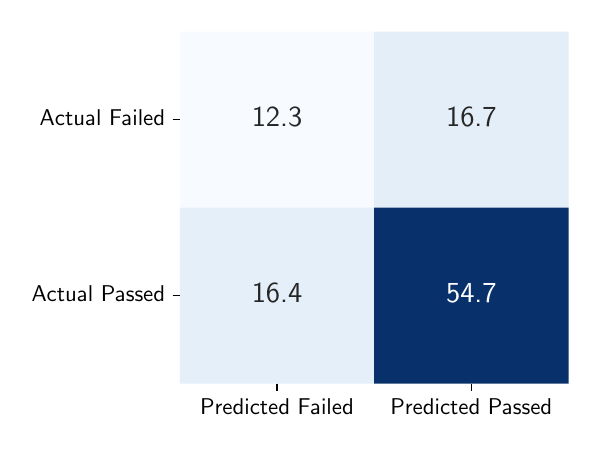}
		\caption{Molmo-7B-D-0924 (baseline)}
		\label{fig:exp_res_confusion_matrix_molmo}
	\end{subfigure}
	\begin{subfigure}{0.44\textwidth}
		\centering
		\includegraphics[width=\linewidth]{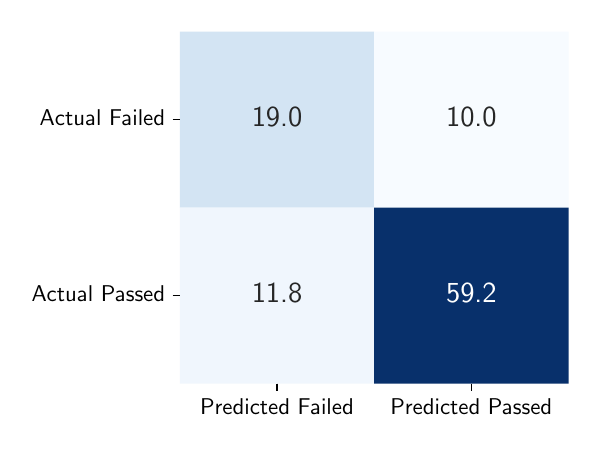}
		\caption{ELAM-7B}
		\label{fig:exp_res_confusion_matrix_elam}
	\end{subfigure}
	\caption{Confusion matrices for the \textit{Expected Results} evaluation classification in AutomotiveUI-Bench-4K, with values normalized to percentages.}
	\label{fig:exp_res_confusion_matrices}
\end{figure*}

The performance of various baseline and fine-tuned models is presented in \cref{tab:hmi_bench_results}.
\ac{elam} demonstrates a notable improvement in localization as indicated by $TA_{vg}$ (+16.3\%) and $ER_{vg}$ (+6.1\%) compared to the baseline model. 
Specifically for evaluation, the fine-tuned model achieved higher accuracy $ER_{evl}$  (+11.3\%), precision (+8.9\%), and recall (+6.4\%) (\cref{fig:exp_res_confusion_matrices}). 
This enhancement is primarily driven by a significant reduction in both false positives and false negatives, indicating fine-tuning effectively improved the model's ability to correctly identify passed tests while simultaneously reducing misclassifications of failed tests as passed, and vice versa.

The findings suggest that models within the 7B parameter range exhibit no significant performance degradation when processing images with predominantly German text.
A significant challenge for most baseline models was the generation of consistently parseable output.
For example, InternVL2.5 exhibited good evaluation performance.
However, it rarely generated bounding box coordinates.
TinyClick and UGround were evaluated exclusively on their localization performance, as they lack evaluative capabilities.
To explore the possibility of faster inference through task distribution across models, TinyClick was selected for \ac{lora}-fine-tuning, given its small size of only 0.27 billion parameters.
This approach (\textit{LAM-270M}), however, did not produce performance on par with Molmo-7B, as shown in \cref{tab:hmi_bench_results}.

A notable benefit of  \ac{elam} is its accessibility for deployment. Unlike many larger state-of-the-art \acp{vlm}, the 7 billion parameter \ac{elam} can be deployed on a current consumer-grade NVIDIA GPU with at least 24 GB VRAM.
The invoke time has been recorded as an average of 2.4 seconds for test actions and 3.4 seconds for expected results when running on an NVIDIA GeForce RTX 4090 GPU, using the default Hugging Face Transformers backend and full resolution images of AutomotiveUI-Bench-4K.

 To assess \ac{elam}'s potential on the AutomotiveUI-Bench-4K dataset, an evaluation was conducted by an experienced quality assurance engineer. This expert achieved improvements of  +6.9\% $TA_{vg}$, +8,9\% $ER_{vg}$, and +15.0\% $ER_{evl}$ compared to \ac{elam}.  This result provides a high-performance human expert benchmark and suggests that enhanced outcomes can be accomplished through additional training.

This analysis with a human expert was instrumental in identifying inherent limitations within the synthetic dataset itself. The expert's higher performance serves as an important indicator for data-related issues. For instance, the boxes may be insufficiently sized for visual grounding, or screen elements may not be sufficiently described, highlighting potential areas for dataset improvement. A detailed analysis is conducted in \cref{sec:visual_error_analysis}.

\begin{table*}[ht]
	\centering
	\small
	\caption{Evaluation on AutomotiveUI-Bench-4K. $TA_{vg}$ and $ER_{vg}$ represent visual grounding accuracies for \textit{Test Actions} and \textit{Expected Results}, respectively. $ER_{evl}$ denotes the classification accuracy (passed/failed) for \textit{Expected Results}. Language specific metrics are indicated by the superscipts $^{EN}$ and $^{DE}$ for English and German UIs. All values are accuracies in percent.}
	\label{tab:hmi_bench_results}
	\begin{threeparttable}
		\begin{tabular*}{\linewidth}{@{\extracolsep{\fill}}l@{}c@{}c@{}c@{}c@{}c@{}c@{}c@{}c@{}c@{}}
			\toprule
			Model &$\downarrow\! TA_{vg}$ & $TA_{vg}^{DE}$ & $TA_{vg}^{EN}$ & $ER_{vg}$ & $ER_{vg}^{DE}$ & $ER_{vg}^{EN}$ & $ER_{evl}$ & $ER_{evl}^{DE}$ & $ER_{evl}^{EN}$ \\
			\midrule
			InternVL2.5-8B\tnote{*} \cite{chen2025internvl25} & 26.6 & 26.1 & 27.0 & 5.7 & 6.3 & 5.1 & 64.8 & 60.4 & 69.0 \\
			TinyClick \cite{pawlowski2024tinyclick} & 61.0 & 54.6 & 67.8 & 53.3 & 47.3 & 59.2 & - & - & - \\
			UGround-V1-7B (Qwen2-VL) \cite{gou2025uground} & 69.4 & 68.9 & 69.9 & 55.0 & 54.4 & 55.7 & - & - & - \\
			Molmo-7B-D-0924 \cite{deitke2024molmo} & 71.3 & 70.9 & 71.8 & 71.4 & 69.8 & 72.9 & 66.9 & 67.8 & 66.0 \\
			LAM-270M (TinyClick) & 73.9 & 66.9 & 81.1 & 59.9 & 54.6 & 65.1 & - & - & - \\
			ELAM-7B (Molmo) & \textbf{87.6} & \textbf{87.5} & \textbf{87.6} & \textbf{77.5} & \textbf{77.0} & \textbf{77.9} & \textbf{78.2} & \textbf{78.5} & \textbf{77.8} \\ 
			Human Domain Expert & 94.5 & 94.5 & 94.5 & 86.4 & 87.3 & 85.5 & 93.2 & 93.7 & 92.8 \\
			\bottomrule
		\end{tabular*}
		\begin{tablenotes}
			\small
			\item[*] Since the InternVL2.5 series model natively supports bounding boxes, these were used for evaluation instead of points.
		\end{tablenotes}
	\end{threeparttable}
\end{table*}

\subsection{Evaluating Generalizability with ScreenSpot}
To determine if the fine-tuning resulted in overfitting to the automotive UI domain, \ac{elam}-7B was applied to the ScreenSpot dataset \cite{cheng2024seeclick} and compared to \textit{SeeClick} \cite{cheng2024seeclick}, \textit{ShowUI} \cite{lin2024showui}, \textit{OS-Atlas-Base-7B} \cite{wu2024atlas}, \textit{Molmo-7B-D-0924 \slash Molmo-72B-0924} \cite{deitke2024molmo}, and Qwen2-VL-based \textit{UGround-V1-7B} \cite{gou2025uground}, as shown in \cref{tab:screenspot_results}.
The fine-tuning procedure was found not to impair the generalizability of Molmo.
The average score per category was improved by 5.6\% compared to its baseline (\textit{Molmo-7B-D-0924}) and by 2.2\% compared to \textit{Molmo-72B-0924}.

\begin{table*}[ht!]
	\centering
	\small
	\caption{$TA_{vg}^{(EN)}$ results on ScreenSpot for selected models taken from \cite{gou2025ugroundgithub}}
	\label{tab:screenspot_results}
	\begin{tabular}{lcp{1cm}p{1cm}p{1cm}p{1cm}p{1cm}p{1cm}} 
		\toprule
		ScreenSpot & $\downarrow$Avg & Mob.-Text & Mob.-Icon & Desk.-Text & Desk.-Icon & Web-Text & Web-Icon \\
		\midrule
		SeeClick \cite{cheng2024seeclick} & 53.4 &  78.0 & 52.0 & 72.2 & 30.0 & 55.7 & 32.5 \\ 
		ShowUI-2B \cite{lin2024showui} & 75.1 &92.3&75.5&76.3&61.1&81.7&63.6\\ 
		Molmo-7B-D-0924 \cite{deitke2024molmo} & 75.2 & 85.4 & 69.0 & 79.4 & 70.7 & 81.3 & 65.5 \\ 
		UGround-V1-2B (Qwen2-VL) \cite{gou2025uground} & 77.7 & 89.4 & 72.0 & 88.7 & 65.7 & 81.3 & 68.9 \\ 
		Molmo-72B-0924 \cite{deitke2024molmo} & 78.6 &  92.7 & 79.5 & 86.1 & 64.3 & 83.0 & 66.0 \\ 
		\textbf{ELAM-7B} (Molmo) & 80.8 & \textbf{94.5} & 79.5 & 89.2 & 70.7 & 85.7 & 65.0 \\ 
		OS-Atlas-Base-7B \cite{wu2024atlas}& 81.0 & 93.0 & 72.9 & 91.8 & 62.9 & 90.9 & 74.3 \\
		UGround-V1-7B (Qwen2-VL) \cite{gou2025uground} & \textbf{86.3} & 93.0 & \textbf{79.9} & \textbf{93.8} & \textbf{76.4} &\textbf{90.9} & \textbf{84.0} \\ 
		\bottomrule
	\end{tabular}
\end{table*}

\subsection{Ablation Study} \label{sec:ablation}

Ablation studies were performed to evaluate the contributions of simple reasoning and rephrasing modules integrated into the synthetic data generation pipeline.
Training runs were conducted using distinct preprocessing strategies, the configurations of which are summarized in \cref{tab:ablation_reasoning_rephrasing}.
The prompts from \cref{sec:elam-prompt-templates} were used for experiments that incorporate reasoning, whereas the phrase ``Think step by step, conclude'' was replaced by ``Determine'' for runs that did not utilize reasoning.
A \ac{lora}-rank of 64 was chosen to fine-tune all linear layers for the ablation experiments.
It was demonstrated that the inclusion of reasoning significantly enhances performance, as evidenced by improvements in both visual grounding ($TA_{vg}, ER_{vg}$) and evaluation ($ER_{evl}$) accuracy.
The critical role of reasoning for robust generalization to real-world data in potential future deployments is highlighted by this observation.

To optimize the \ac{lora} configuration, a separate ablation study focusing on the rank parameter $r$ was conducted, the results of which are presented in \cref{tab:ablation_lora}.
The data was preprocessed utilizing reasoning and rephrasing.
This analysis determined that a rank of 64 yielded optimal performance.
In all \ac{lora}-based experiments, the alpha parameter was set to $r$, and the \ac{lora}-dropout rate was fixed at $0.05$.

\noindent
\begin{minipage}[ht]{0.48\textwidth}
	\centering
	\captionof{table}{Prompting ablations}
	\label{tab:ablation_reasoning_rephrasing}
	\begin{tabular}{lccc}
		\toprule
		Settings & $TA_{vg}$ & $ER_{vg}$ & $ER_{evl}$\\
		\midrule
		Baseline & 81.7 & 73.5 & 73.5 \\ 
		Rephrasing & 83.9 & 73.7 & 71.1 \\ 
		Reasoning & 86.4 & 76.2 & 77.0 \\ 
		Reas.+Rephr. &\textbf{87.4}&\textbf{77.2}&\textbf{78.6}\\ 
		\bottomrule
	\end{tabular}
	
\end{minipage}
\hfill 
\begin{minipage}[ht]{0.48\textwidth} 
	\centering
	\captionof{table}{LoRa ablations}
	\label{tab:ablation_lora}
	\begin{tabular}{lccc}
		\toprule
		Rank & $TA_{vg}$ & $ER_{vg}$ & $ER_{evl}$\\
		\midrule
		16 & 84.6 & 76.1 & 77.7 \\ 
		32 & 85.0 & 76.4 & 77.1 \\ 
		64 & \textbf{87.4}  & \textbf{77.2} & \textbf{78.6}\\ 
		128 & 87.1  & \textbf{77.2} & 77.6 \\ 
		\bottomrule
	\end{tabular}
\end{minipage}

\subsection{Analysis of Utterance Embedding Space vs. \ac{vlm} Performance}

To investigate the relationship between the semantic representation of user utterances and \ac{vlm} performance, particularly within automotive subdomains, two distinct task types were analyzed: \textit{Test Actions} and \textit{Expected Results}, as outlined in \cref{sec:novel_automotive_ui_validation_dataset}.
For both types, text embeddings of the utterances were generated using the \textit{nomic-embed-text-v1.5}\footnote{\textit{nomic-embed-text-v1.5} is an improved variant of the \textit{nomic-embed-text-v1}\cite{nussbaum2024nomic} model.
It utilizes Matryoshka Representation Learning for flexible dimensionality reduction with minimal loss while supporting up to 8,192 tokens.
The model is optimized for search, clustering, and classification in production.} model.
Their structure was then visualized using t-SNE dimensionality reduction \cite{vandermaaten08a}.
Unsupervised k-Means clustering ($k=8$) was applied to the high-dimensional embeddings to identify potential thematic groups.
Cluster labels were generated post-hoc via a \ac{llm} (\textit{gpt-4o-2024-11-20}) interpretation of sampled utterances.
The t-SNE projections were enhanced with a heatmap that visualizes the local failure rate for each designated target for \textit{Test Actions} and both evaluation and grounding for \textit{Expected Results}.
This failure rate was calculated per grid cell based on evaluation results.
Red indicates high failure density, and white indicates low failure density.
This approach facilitates the visualization of how \ac{vlm} performance, for both the base Molmo-7B and \ac{elam}-7B models, varies across the semantic landscape defined by the text embeddings.

\subsubsection{\textit{Test Action} Utterances: Grounding Performance}

\begin{figure}[htbp]
	\centering
	\includegraphics[width=0.75\linewidth]{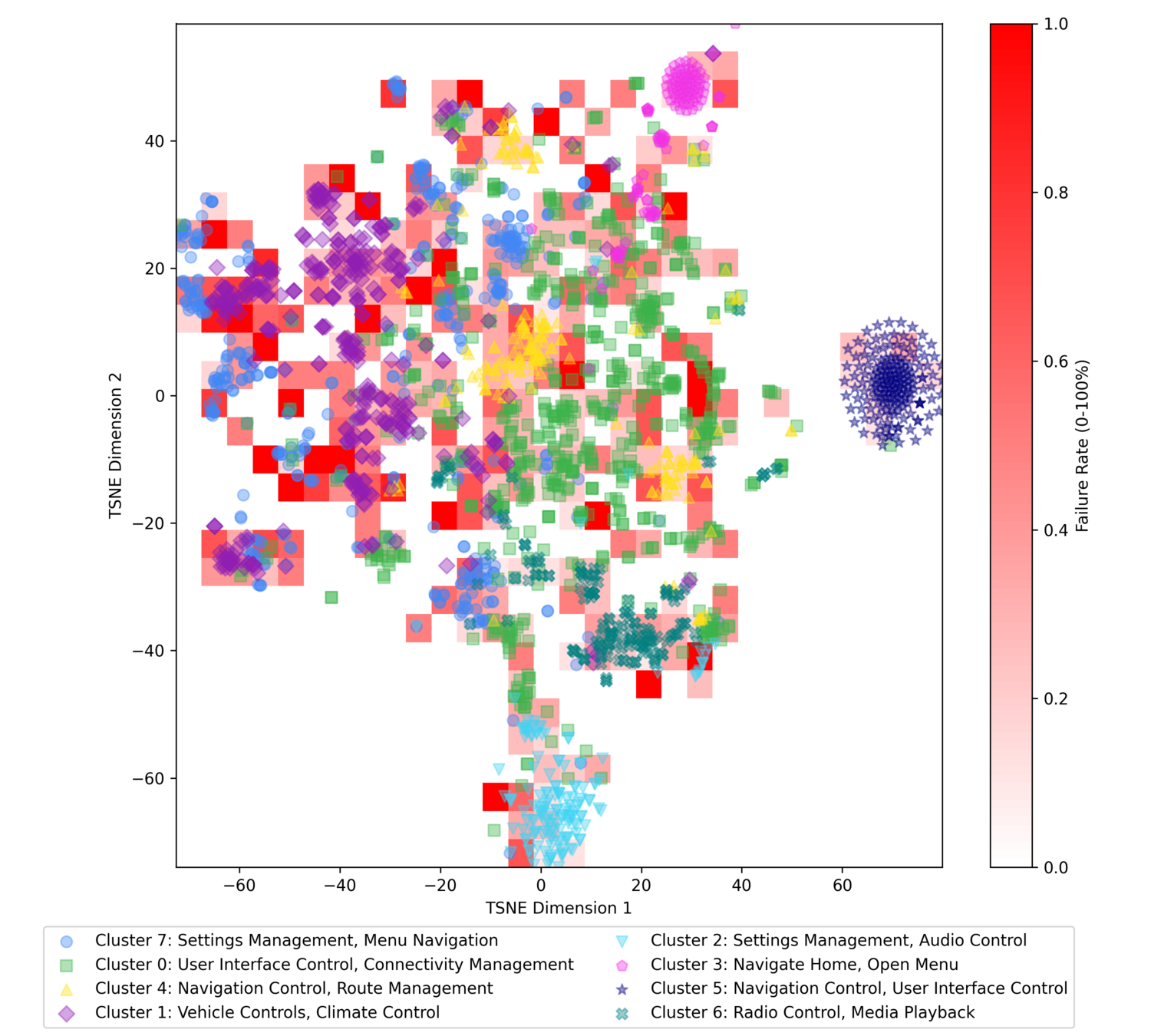}
	\Description{This image is a scatter plot that uses different shapes and colors to represent different categories of car functions, and it overlays red shading to show the failure rate associated with those functions. The two axes correspond to TSNE dimension 1 and 2.
	K-Means returns 8 clusters:
	Cluster 0 (Light Green Square): "User Interface Control, Connectivity Management".
	Cluster 1 (Purple Diamond): "Vehicle Controls, Climate Control".
	Cluster 2 (Light Blue Downward Triangle): "Settings Management, Audio Control".
	Cluster 3 (Pink Star): "Navigate Home, Open Menu".
	Cluster 4 (Yellow Triangle): "Navigation Control, Route Management".
	Cluster 5 (Dark Blue Star): "Navigation Control, User Interface Control".
	Cluster 6 (Teal Cross): "Radio Control, Media Playback".
	Cluster 7 (Light Blue Circle): Imagine many small, light blue circles. These represent "Settings Management, Menu Navigation".
	}
	\caption{t-SNE plot of the base Molmo-7B model for the \textit{Test Action} task}
	\label{fig:testmolmo-7b-d-base-092420250305-153817-9999testactiontsneheatmap}
\end{figure}

\paragraph{Baseline Molmo-7B-D-0924:}
Clusters with \ac{llm}-suggested labels indicating semantic overlap (e.g., 'Navigation Control' across multiple clusters) are observed in the t-SNE plot (\cref{fig:testmolmo-7b-d-base-092420250305-153817-9999testactiontsneheatmap}) of \textit{Test Action} embeddings for the base model. The grounding failure heatmap indicates that errors are distributed across the embedding space, without perfectly aligning with cluster boundaries. Notable high-failure regions overlap with clusters tentatively labeled \textit{Climate Control, Audio and Display Settings} and \textit{User Interface Control, Device Settings}. A distinct cluster associated with \textit{Navigation Control, User Interface Control} resides in a predominantly low-failure region, which suggests that these utterance types are generally grounded successfully.

\begin{figure}[htbp]
	\centering
	\includegraphics[width=0.75\linewidth]{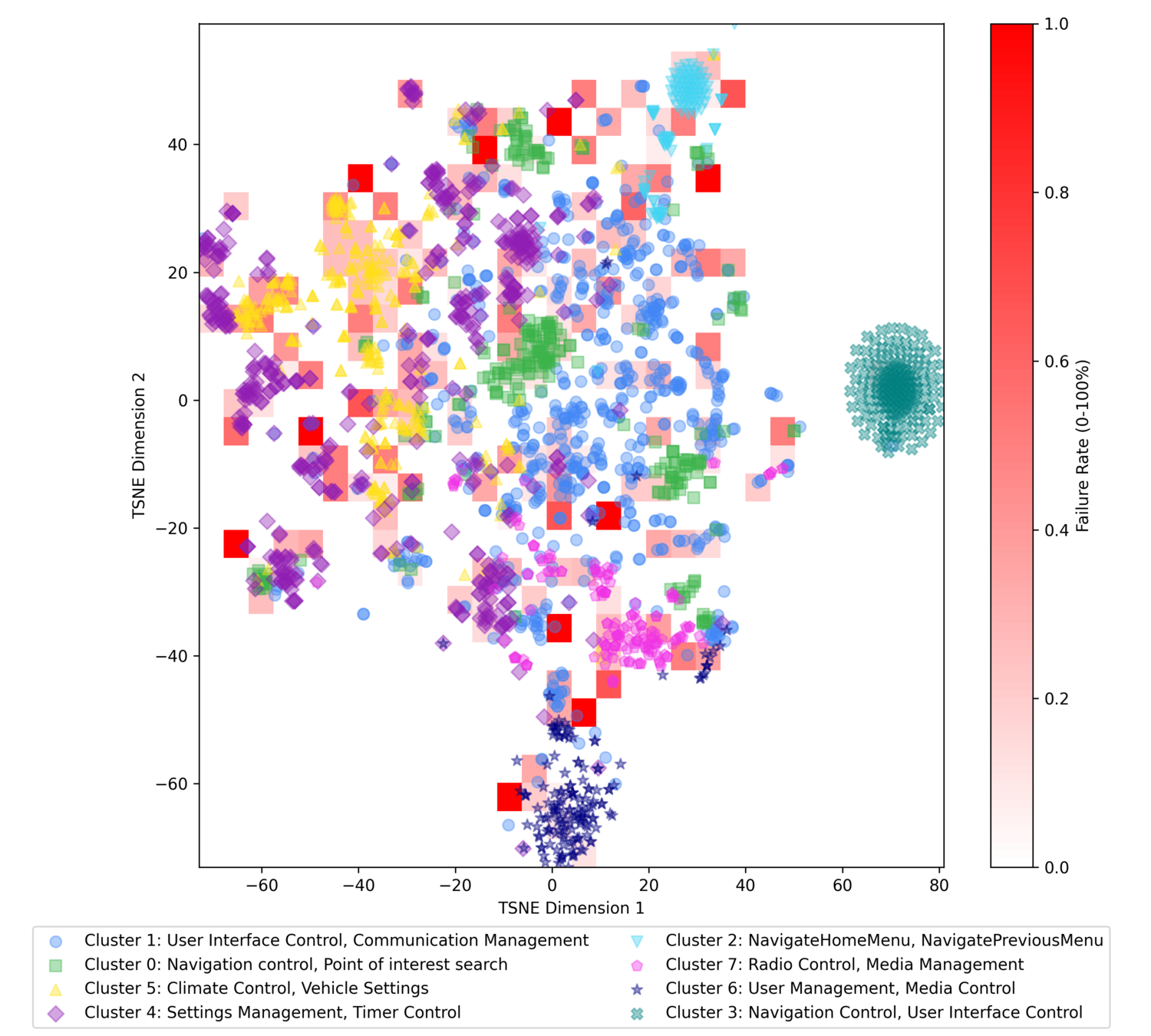}
	\Description{This image is a scatter plot that uses different shapes and colors to represent different categories of car functions, and it overlays red shading to show the failure rate associated with those functions. The two axes correspond to TSNE dimension 1 and 2.
	K-Means returns 8 clusters:
	Cluster 0 (Light Green Square): Navigation Control, Point of interest search
	Cluster 1 (Light Blue Circle): User Interface Control, Communication Management
	Cluster 2 (Light Blue Downward Triangle): NavigateHomeMenu, NavigatePreviousMenu
	Cluster 3 (Teal Cross): Navigation Control, User Interface Control
	Cluster 4 (Purple Diamond): Settings Management, Timer Control
	Cluster 5 (Yellow Triangle): Climate Control, Vehicle Settings
	Cluster 6 (Dark Blue Star): User Management, Media Control
	Cluster 7 (Pink Star): Radio Control, Media Management
	}
	\caption{t-SNE plot of the \ac{elam}-7B model for the \textit{Test Action} task}
	\label{fig:testmolmo-7b-d-0924-0x001520250321-081639-50000testactiontsneheatmap}
\end{figure}

\paragraph{\ac{elam}-7B:}
Utilizing the same text embeddings, the fine-tuned model's performance is visualized through its distinct failure heatmap (\cref{fig:testmolmo-7b-d-0924-0x001520250321-081639-50000testactiontsneheatmap}).
When comparing \cref{fig:testmolmo-7b-d-0924-0x001520250321-081639-50000testactiontsneheatmap} to \cref{fig:testmolmo-7b-d-base-092420250305-153817-9999testactiontsneheatmap}, a noticeable reduction in both the extent and intensity of high-failure regions across the embedding space can be observed, supporting the results summarized in \cref{tab:hmi_bench_results}.
Several clusters show significant improvement.
For instance, the most distinct \textit{Cluster 3: Navigation Control, User Interface Control} exhibited a failure rate close to zero.
The dispersed \textit{Cluster 1: User Interface Control, Communication Management}, \textit{Cluster 4: Climate Control, Vehicle Settings} and \textit{Cluster 5: Climate Control, Vehicle Settings} now predominantly reside in low-failure regions after fine-tuning, indicating improved reliability in handling these automotive domain-specific utterance types.
However, some high-failure samples still persist at the boundaries of these clusters, suggesting that certain semantic regions or utterance formulations remain challenging even after fine-tuning.

\subsubsection{\textit{Expected Result} Utterances: Evaluation and Grounding Performance}

\begin{figure}[htbp]
	\centering
	\begin{subfigure}[b]{0.49\linewidth}
		\centering
		\includegraphics[width=\linewidth]{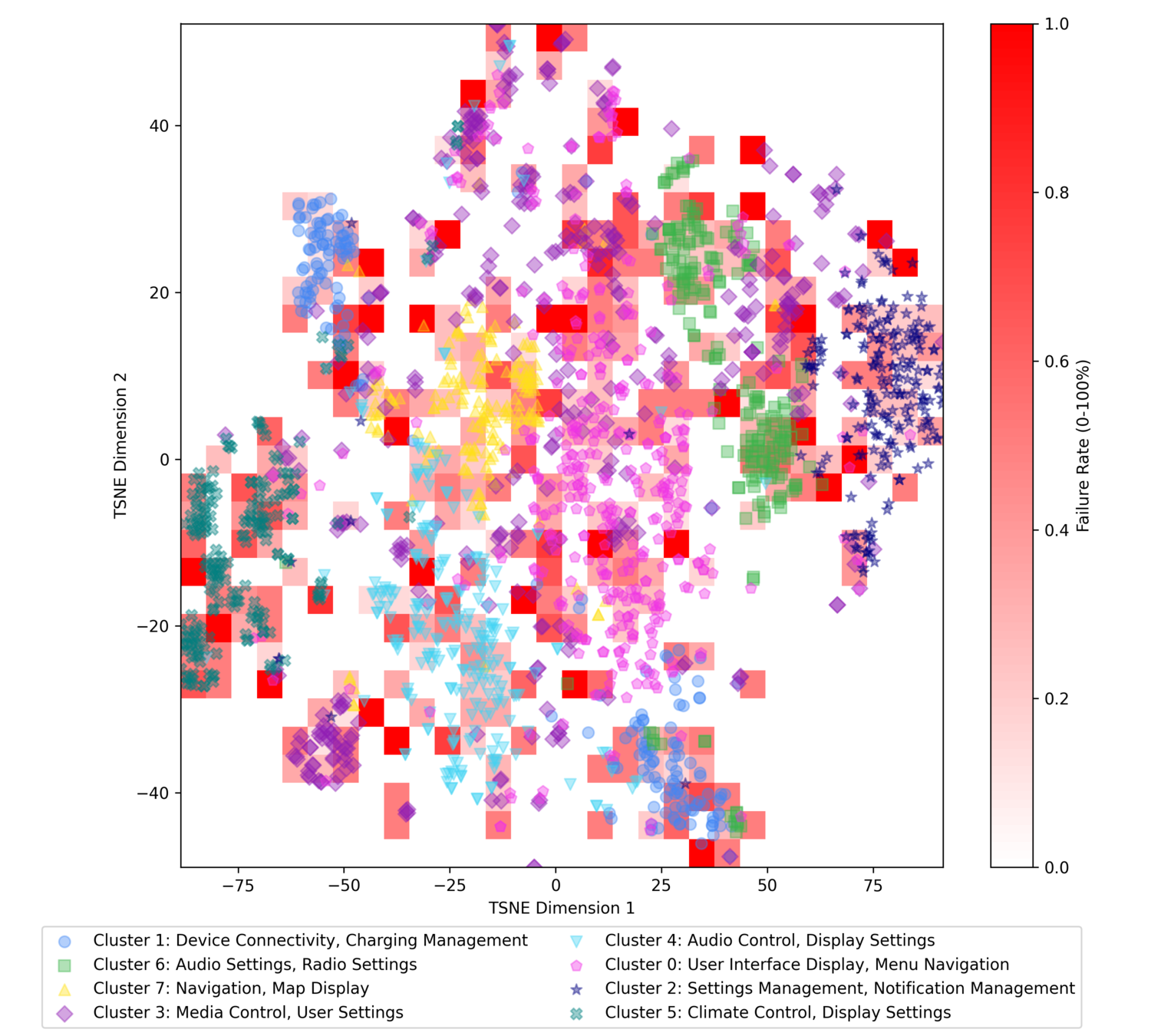}
		\Description{This image is a scatter plot that uses different shapes and colors to represent different categories of car functions, and it overlays red shading to show the failure rate associated with those functions. The two axes correspond to TSNE dimension 1 and 2.
		Cluster 0 (Pink Star): User Interface Display, Menu Navigation
		Cluster 1 (Light Blue Downward Triangle): Device Connectivity, Charging Management
		Cluster 2 (Dark Blue Star): Settings Management, Notification Management
		Cluster 3 (Purple Diamond): Media Control, User Settings
		Cluster 4 (Light Blue Circle): Audio Control, Display Settings
		Cluster 5 (Teal Cross): Climate Control, Display Settings
		Cluster 6 (Light Green Square): Audio Settings, Radio Settings
		Cluster 7 (Yellow Triangle): Navigation, Map Display
		}
		\caption{}
		\label{fig:tsne_base_exp_res}
	\end{subfigure}
	\begin{subfigure}[b]{0.49\linewidth}
		\centering
		\includegraphics[width=\linewidth]{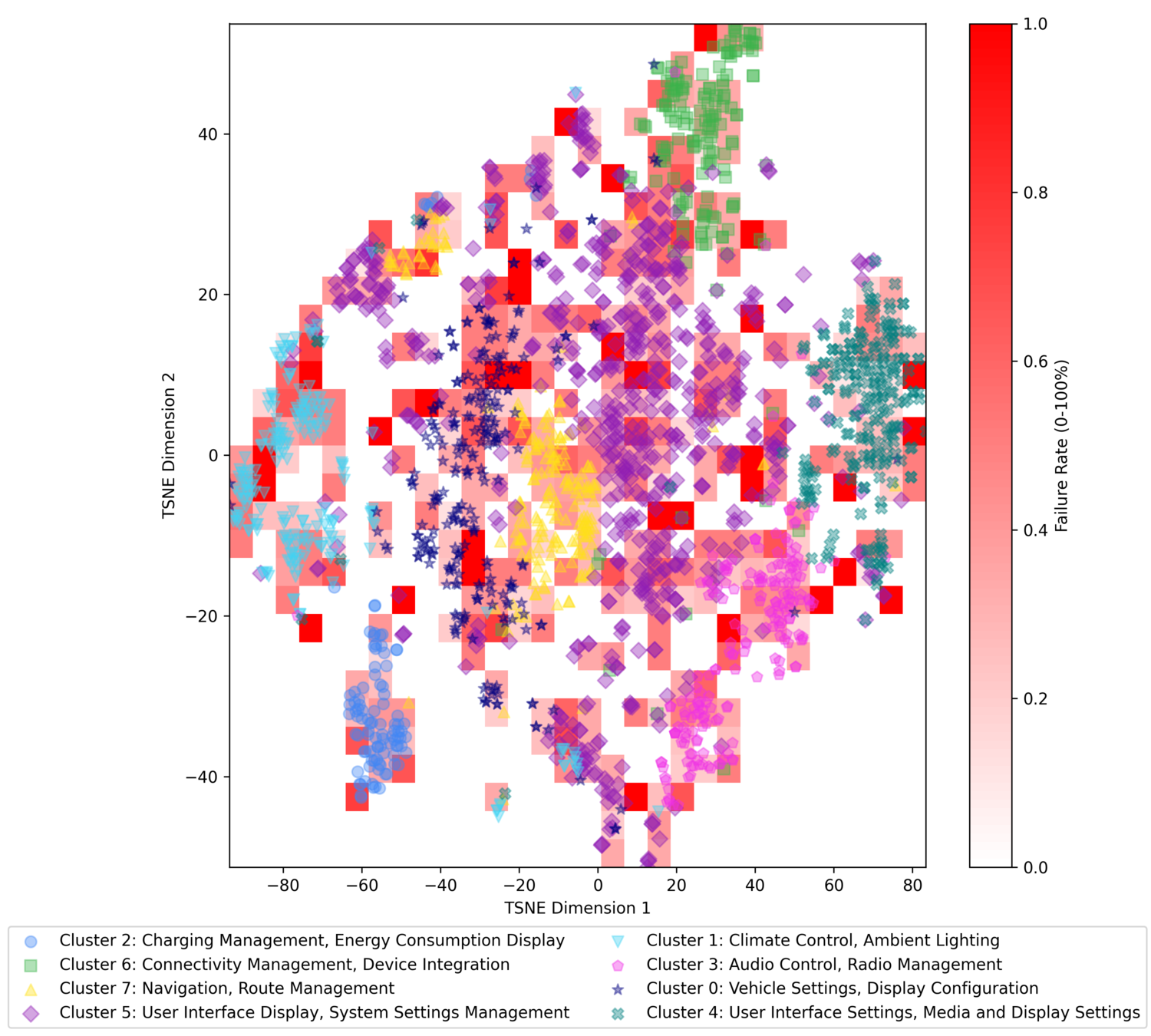}
		\Description{This image is a scatter plot that uses different shapes and colors to represent different categories of car functions, and it overlays red shading to show the failure rate associated with those functions. The two axes correspond to TSNE dimension 1 and 2.
		Cluster 0 (Teal Cross): Vehicle Settings, Display Configuration
		Cluster 1 (Light Blue Downward Triangle): Climate Control, Ambient Lighting
		Cluster 2 (Light Blue Circle): Charging Management, Energy Consumption Display
		Cluster 3 (Pink Star): Audio Control, Radio Management
		Cluster 4 (Light Green Crosshair): User Interface Settings, Media and Display Settings
		Cluster 5 (Purple Diamond): User Interface Display, System Settings Management
		Cluster 6 (Light Green Square): Connectivity Management, Device Integration
		Cluster 7 (Yellow Triangle): Navigation, Route Management
		}
		\caption{}
		\label{fig:tsne_base_exp_res_conclusion}
	\end{subfigure}
	\caption{t-SNE visualizations of the base Molmo-7B model for two \textit{Expected Result} evaluation tasks: (a) \textit{Visual Grounding} and (b) \textit{Evaluation}.}
	\label{fig:tsne_base_combined}
\end{figure}

\paragraph{Baseline Molmo-7B-D-0924:}
The \textit{Expected Result} utterances, representing a more complex task involving visual grounding and subsequent evaluation, are also observed to form clusters in the embedding space with themes such as ``Settings'', ``Navigation'', and ``Display''.
The Molmo-7B failure heatmap (\cref{fig:tsne_base_combined}) for this task exhibits widespread and often intense high-failure regions, particularly concentrated in areas associated with \textit{User Interface Settings, Search Functionality} and \textit{Audio Control, Radio Management}.

\begin{figure}[htbp]
	\centering
	\begin{subfigure}[b]{0.48\linewidth}
		\centering
		\includegraphics[width=\linewidth]{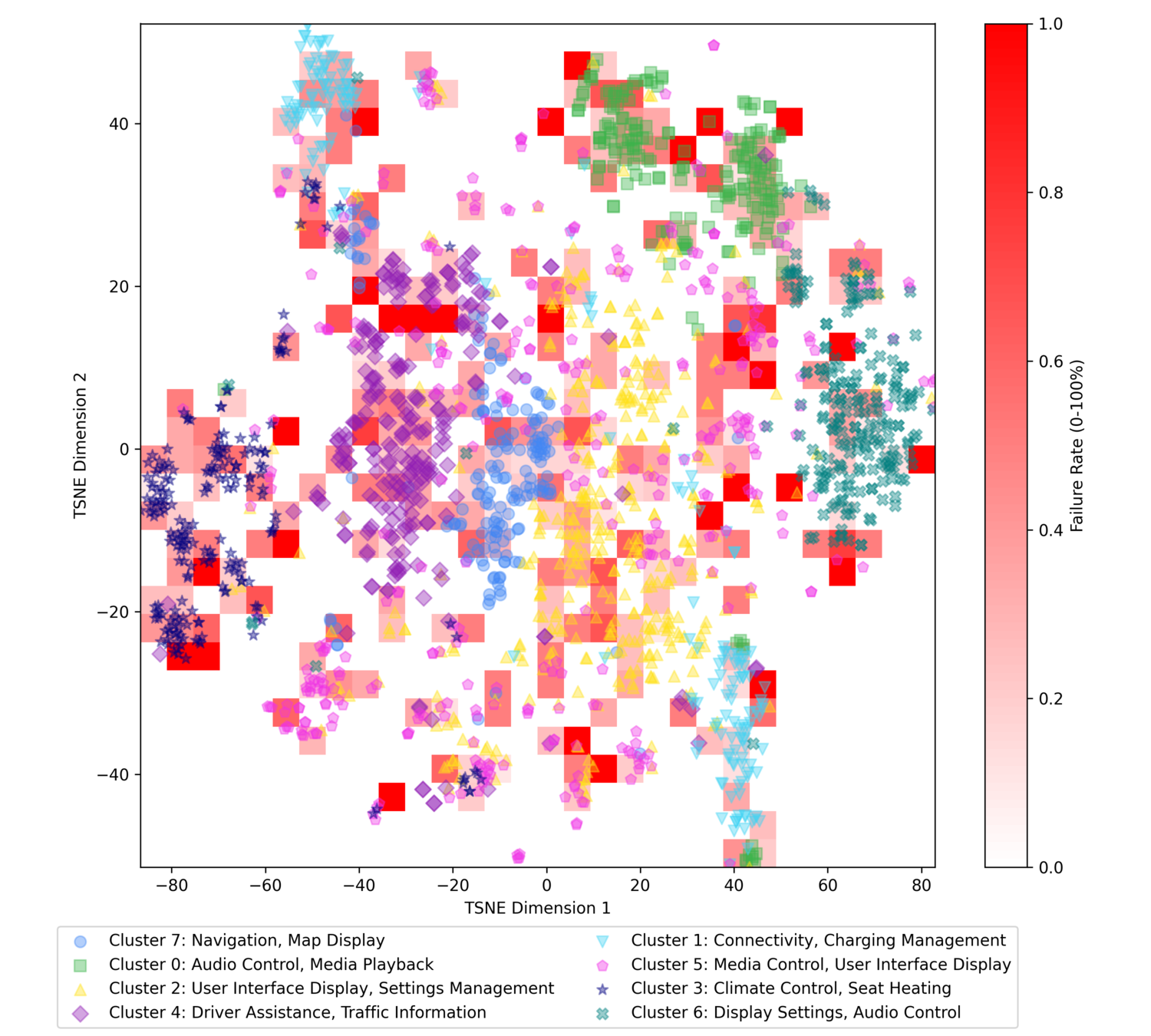}
		\Description{This image is a scatter plot that uses different shapes and colors to represent different categories of car functions, and it overlays red shading to show the failure rate associated with those functions. The two axes correspond to TSNE dimension 1 and 2.
		Cluster 0 (Light Green Square): Audio Control, Media Playback
		Cluster 1 (Light Blue Downward Triangle): Connectivity, Charging Management
		Cluster 2 (Yellow Triangle): User Interface Display, Settings Management
		Cluster 3 (Teal Cross): Climate Control, Seat Heating
		Cluster 4 (Purple Diamond): Driver Assistance, Traffic Information
		Cluster 5 (Pink Star): Media Control, User Interface Display
		Cluster 6 (Light Green Crosshair): Display Settings, Audio Control
		Cluster 7 (Light Blue Circle): Navigation, Map Display
		}
		\caption{}
		\label{fig:tsne_elam_exp_res}
	\end{subfigure}
	\begin{subfigure}[b]{0.48\linewidth}
		\centering
		\includegraphics[width=\linewidth]{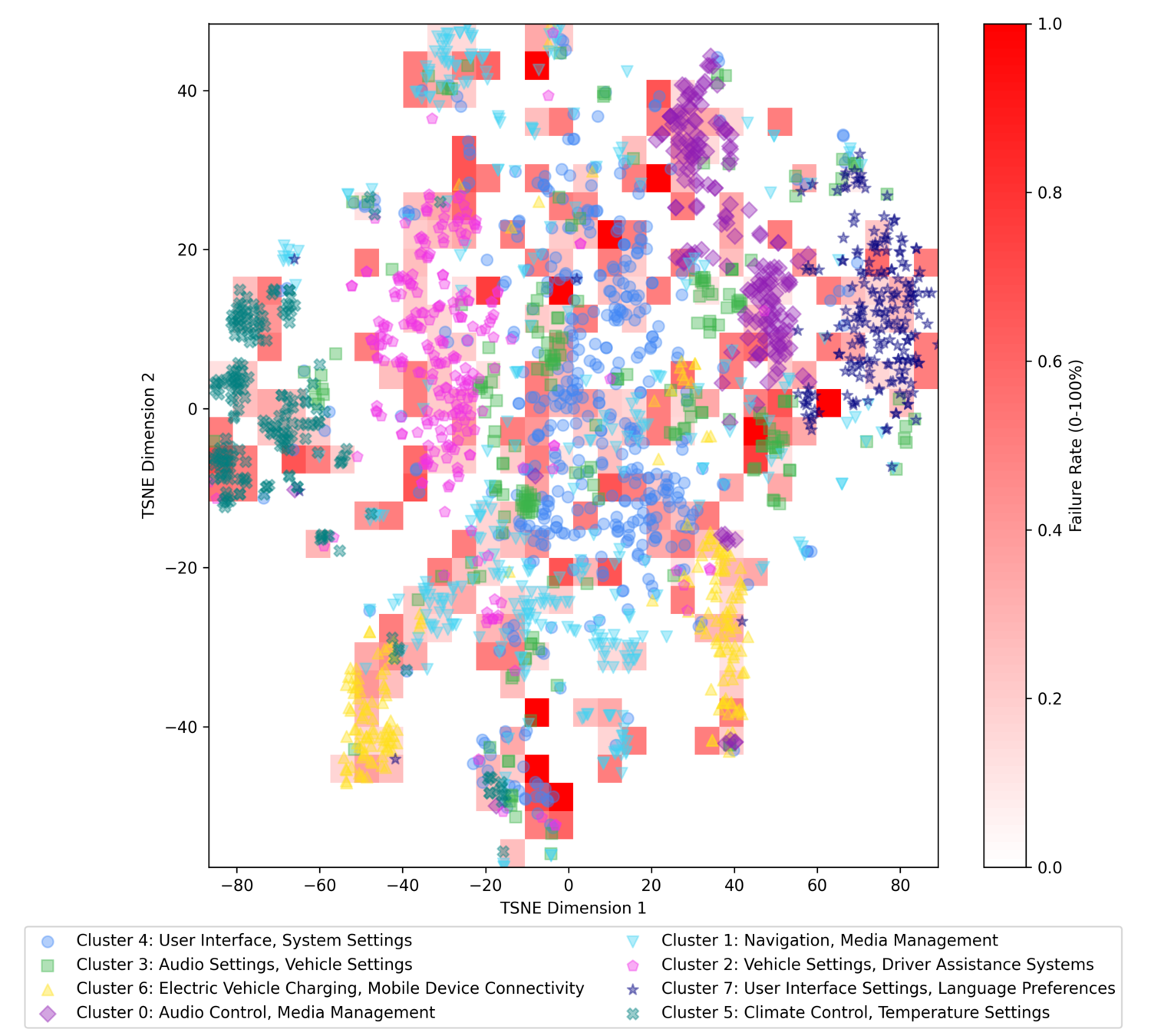}
		\Description{This image is a scatter plot that uses different shapes and colors to represent different categories of car functions, and it overlays red shading to show the failure rate associated with those functions. The two axes correspond to TSNE dimension 1 and 2.
		Cluster 0 (Purple Diamond): Audio Control, Media Management
		Cluster 1 (Light Blue Downward Triangle): Navigation, Media Management
		Cluster 2 (Pink Star): Vehicle Settings, Driver Assistance Systems
		Cluster 3 (Light Green Square): Audio Settings, Vehicle Settings
		Cluster 4 (Light Blue Circle): User Interface, System Settings
		Cluster 5 (Teal Cross): Climate Control, Temperature Settings
		Cluster 6 (Yellow Triangle): Electric Vehicle Charging, Mobile Device Connectivity
		Cluster 7 (Dark Blue Star): User Interface Settings, Language Preferences
		}
		\caption{}
		\label{fig:tsne_elam_exp_res_conclusion}
	\end{subfigure}
	\caption{t-SNE visualizations of the \ac{elam}-7B model for two \textit{Expected Result} evaluation tasks: (a) \textit{Visual Grounding} and (b) \textit{Evaluation}.}
	\label{fig:tsne_elam_combined}
\end{figure}

\paragraph{\ac{elam}-7B:}
The failure heatmap (\cref{fig:tsne_elam_combined}) for the fine-tuned model on the \textit{Expected Result} tasks indicates that high-failure regions persist across many parts of the embedding space. While a visual comparison with the base Molmo model may suggest a subtle reduction in the overall failure density, the improvement appears less pronounced than that observed for the \textit{Test Action} task.

Fine-tuning yielded only modest improvements for the more complex \textit{Expected Result} evaluation and grounding task. The widespread persistence of high-failure regions indicates that accurately interpreting and verifying diverse state descriptions based on visual evidence remains a significant challenge.

\subsection{Visual Error Analysis} \label{sec:visual_error_analysis}
\subsubsection{Visual Error Analysis: Test Actions}
A detailed analysis of the \ac{elam}'s failures in the AutomotiveUI-Bench-4K evaluation was conducted to identify the areas where the model continues to demonstrate deficiencies. To begin with, the \textit{Test Actions} were examined. A comprehensive manual review and classification of all error cases was conducted.
Examples and explanations for all identified categories can be found in the appendix in \cref{sec:examples_of_visual_errors_ta}. \cref{tab:elam_ta_error_categories} summarizes the distribution of error categories for \textit{Test Actions}.

\begin{samepage}
	The following categories were identified: 
	\begin{enumerate}[topsep=0pt, itemsep=0pt, parsep=0pt]
		\item The bounding boxes utilized for the purpose of evaluating the result were of insufficient size.
		\item The element to be determined was not adequately described.
		\item The subject of interest was a specialized automotive icon.
		\item The provided description was adequate; however, the presence of two similar elements resulted in confusion.
		\item The model proved incapable of establishing a connection between two elements.
		\item The model demonstrated confusion between driver side and passenger side, left and right, or increase and decrease.
		\item The model was incapable of counting.
		\item Not assignable to any category.
	\end{enumerate}
\end{samepage}

\begin{table*}[ht!]
	\centering
	\small
	\caption{Analysis of error categories based on \textit{Test Actions} in AutomotiveUI-Bench-4K.}
	\label{tab:elam_ta_error_categories}
	\begin{tabular}{cp{9cm}c}
		\toprule
		\textbf{Category \#} & \textbf{Category Description} & \textbf{Error Share in \%} \\
		\midrule
		\textbf{1.} & Box too small & 17.0 \\
		\textbf{2.} & Ambiguous or poor description of the test action & 23.7 \\
		\textbf{3.} & Special icons from the vehicle domain & 19.7 \\
		\textbf{4.} & Clear description, but very similar elements confused the model & 11.5 \\
		\textbf{5.} & Connection between elements could not be established & 6.8 \\
		\textbf{6.} & Driver/passenger, left/right, up/down distinctions & 4.8 \\
		\textbf{7.} & Elements must be counted & 3.1 \\
		\textbf{8.} & Miscellaneous & 13.4 \\
		\bottomrule
	\end{tabular}
\end{table*}

\noindent
The findings underscore the importance of ensuring clear and precise phrasing in the test utterances used by developers and testers, as this practice can mitigate a range of potential issues. Analysis also showed that the system's recognition of vehicle-specific icons, in particular, remains suboptimal, suggesting that performance would benefit from additional training data. A more fundamental challenge identified is the system's difficulty in differentiating between highly similar elements and establishing connections between them. The creation of specialized training datasets to address these specific scenarios represents a promising avenue for future research.

\subsubsection{Visual Error Analysis: Expected Results}
An analysis of the \ac{elam}'s failures of grounding and evaluating \textit{Expected Results} in the AutomotiveUI-Bench-4K evaluation was conducted in the same manner as the examination of the \textit{Test Actions}. The errors were manually reviewed and categorized.
The identified categories were similar to those found in the examination of the \textit{Test Actions}, though some differences were noted.
\cref{tab:elam_er_error_categories} summarizes the distribution of error categories for \textit{Expected Results}. Examples and explanations for all identified categories can be found in the appendix in \cref{sec:examples_of_visual_errors_er}.

\begin{samepage}
	The following categories were identified: 
	\begin{enumerate}[topsep=0pt, itemsep=0pt, parsep=0pt]
		\item The bounding boxes utilized for the purpose of evaluating the result were of insufficient size.
		\item Multiple areas display the expected result.
		\item The model is asked to check if options of a certain menu are visible. However the box was not drawn around the menu option, but around the menu heading instead. This could be considered as a special case of category 1 or 2.
		\item The subject of interest was a specialized automotive icon.
		\item The provided description was adequate, however the presence of two similar elements resulted in confusion.
		\item Not assignable to any category.
	\end{enumerate}
\end{samepage}

\begin{table*}[ht!]
	\centering
	\small
	\caption{Analysis of error categories based on \textit{Expected Results} in AutomotiveUI-Bench-4K.}
	\label{tab:elam_er_error_categories}
	\begin{tabular}{cp{9cm}c}
		\toprule
		\textbf{Category \#} & \textbf{Category Description} & \textbf{Error Share in \%} \\
		\midrule
		\textbf{1.} & Box too small & 26.4 \\
		\textbf{2.} & Multiple areas display the expected result & 15.2 \\
		\textbf{3.} & Box not around option, when asked if options are shown & 15.2 \\
		\textbf{4.} & Special icons from the vehicle domain & 8.8 \\
		\textbf{5.} & Clear description, but very similar elements confused the model & 4.8 \\
		\textbf{6.} & Miscellaneous & 29.6 \\
		\bottomrule
	\end{tabular}
\end{table*}

\section{Conclusion}

A clear improvement in localization and evaluation performance is demonstrated using a \ac{lora} fine-tuned model, trained with a synthetic data generation pipeline.
\ac{elam} outperforms its baseline on both AutomotiveUI-Bench-4K and ScreenSpot, highlighting this approach as an effective strategy for enhancing visual grounding capabilities with limited resources.
The model demonstrates a notable improvement in localization accuracy on the AutomotiveUI-Bench-4K dataset, achieving a gain of 16.3\% for \textit{Test Actions} ($TA_{vg}$), 6.1\% for \textit{Expected Results} ($ER_{vg})$, and 11.3\% evaluation conclusion accuracy ($ER_{evl}$) compared to its baseline \textit{Molmo-7B-D-0924}.
These improvements maintain generalizability across diverse UI domains, including desktop, mobile, and web, as evidenced by an 80.8\% (+5.6\%) average accuracy on the ScreenSpot dataset.

The t-SNE visualization of a fixed text embedding space overlaid with task-specific \ac{vlm} failure rates provided valuable insights by decoupling utterance semantics from model performance.
The analysis indicated tangible benefits of fine-tuning for more direct tasks such as the \textit{Test Action} grounding, and suggested that more complex tasks like the \textit{Expected Result} evaluation continue to present significant challenges.
Additionally, the t-SNE plots revealed variations in performance across different sub-domains. The model demonstrated strong performance in grounding \ac{ui} elements commonly found in desktop, mobile, and web environments but exhibited reduced effectiveness in automotive-specific functionalities such as \ac{adas} or climate control, for instance, distinguishing between adjusting the driver's and passenger's temperatures.
This emphasizes the general need for further domain-specific data and training strategies.

While the findings are encouraging, it is important to acknowledge certain limitations of this work.
Firstly, while the synthetic data pipeline proved effective, the inherent gap between synthetic and real-world data introduces a potential limitation.
This restricts the applicability of the findings to entirely unconstrained scenarios.
Further investigation is needed to assess performance on a wider range of real-world datasets beyond AutomotiveUI-Bench-4K, and across more diverse visual grounding tasks.
Errors were identified during a preliminary analysis of the training data, including the misclassification of fundamental \ac{ui} element states (e.g., toggle switches). Additionally, the distribution of use cases exhibited an excessive emphasis on the presence and visibility of control elements (e.g.,\textit{``Element XYZ is visible.''}). This indicates that the number of annotations regarding the status of control elements is insufficient.

As a consequence, exclusive reliance on \ac{elam}-7B to automate \ac{ui} verification is cautioned against. For automotive functions that directly impact passenger safety, a \ac{vlm}-based solution introduces significant safety and ethical concerns, making it an unsuitable final authority for correctness.
Given these limitations, the most responsible approach is to position \ac{vlm}-based verification not as a replacement for, but as a supplement to existing verification pipelines. \acp{vlm} can be highly effective in automating the testing of non-critical infotainment features, visual consistency checks, or identifying minor \ac{ui} bugs. For any functionality related to passenger safety, however, a human-in-the-loop protocol is essential. The \ac{vlm} could be used to flag potential issues for human review, but the final, authoritative sign-off on safety-critical \ac{ui} functionality must always be performed by a human expert using a validated method. This hybrid approach leverages the efficiency of machine learning while upholding the paramount ethical obligation to ensure human safety.

The scope of the evaluation, while demonstrating improvement, primarily focused on quantitative metrics.
A deeper qualitative analysis of the model's performance, especially on automotive-specific parts of automotive systems, would provide richer insights into the nuanced strengths and weaknesses of the approach.

Looking ahead, there are several promising avenues for potential improvements. Future work could focus on developing more advanced and general evaluation methodologies.
These methodologies would integrate linguistic and grammatical analysis to better understand the impact of subtle syntactic and semantic nuances on model performance.
This may include incorporating techniques like dependency parsing and syntactic analysis to provide a more comprehensive picture of the interplay between language and visual grounding.

Moreover, there is significant potential for enhancing the capabilities of the visual encoders within \ac{vlm} architectures. Although current pre-trained encoders are effective, they may overlook certain domain-specific features and high-resolution details. Future improvements might explore the use of multi-encoder strategies, adaptive encoder selection, and more data-efficient fine-tuning techniques to overcome these limitations and achieve more robust and generalized performance across diverse real-world scenarios.

Overall, the findings point toward a holistic strategy for advancing \ac{vlm} technologies.
This strategy refines visual grounding through innovative training pipelines and embraces broader evaluation frameworks and encoder enhancements.
These integrated improvements hold the promise of pushing the boundaries of current models toward greater real-world applicability.

\section*{Acknowledgement}
This work was partially supported by German BMBF within the research project MANNHEIM-KI4BoardNet. (\url{https://www.elektronikforschung.de/projekte/mannheim-ki4boardnet}).

\bibliographystyle{acm}
\bibliography{literature}
\newpage
\appendix
\section{Appendix}

\subsection{Code and Dataset Availability} \label{sec:data_availability}
For reproducibility and broader utility, our resources are openly available.
\href{https://huggingface.co/sparks-solutions/ELAM-7B}{\underbar{ELAM}} and its supporting code, alongside the \href{https://huggingface.co/datasets/sparks-solutions/AutomotiveUI-Bench-4K}{\underbar{AutomotiveUI-Bench-4K}} dataset, are hosted on Hugging Face.
The ELAM repository provides the code and prompting examples necessary to replicate the results detailed in this paper.
Consistent with its fine-tuned base, Molmo, ELAM and its code are released under the Apache License 2.0.
The AutomotiveUI-Bench-4K dataset is distributed under the CC-BY-4.0 license.

\subsection{Prompt Templates for ELAM} \label{sec:elam-prompt-templates}
\begin{table*}[ht]
	\begin{tabular}{l}
		\textbf{Prompt Template - Test Action} \\
		\noindent
		\begin{minipage}{0.98\linewidth}
			\noindent
			\begin{python}
				prompt_template_test_action = f"""\
				Identify and point to the UI element that corresponds to this test action:
				{test_action}.
				"""
			\end{python}
		\end{minipage} \\  
		\textbf{Response Template - Test Action} \\
		\begin{minipage}{0.98\linewidth}
			\begin{python}
				response_template_test_action = f"""\
				{reasoning}
				<point x="{center_x:.1f}" y="{center_y:.1f}" alt="{test_action}">\
				{test_action}</point>
				"""
			\end{python}
		\end{minipage} \\
		\textbf{Prompt Template - Expected Result} \\
		\begin{minipage}{0.98\linewidth}
			\begin{python}
				prompt_template_expected_result = f"""\
				Evaluate this statement about the image:
				'{expectation}'
				Think step by step, conclude whether the evaluation is 'PASSED' or 'FAILED'\
				and point to the UI element that corresponds to this evaluation.
				"""
			\end{python}
		\end{minipage} \\
		\textbf{Response Template - Expected Result} \\
		\begin{minipage}{0.98\linewidth}
			\begin{python}
				response_template_expected_result = f"""\
				{reasoning}
				Conclusion: {evaluation_result}
				<point x="{center_x:.1f}" y="{center_y:.1f}" alt="{expectation}">\
				{expectation} {evaluation_result}</point>
				"""
			\end{python}
		\end{minipage} \\
		
	\end{tabular}
\end{table*}

\newpage
\subsection{Prompts in Synthetic Data Pipeline}
\label{sec:synthetic_data_gen_prompts_gemini}
\lstdefinestyle{myPromtStyle}{
    basicstyle=\small,
    breaklines=true,
    frame=single,
    showspaces=false,
    showstringspaces=false
}
\subsubsection{Test Action Template}
\label{promt:synt_test_action}
\begin{lstlisting}[style=myPromtStyle]
Write test actions for the UI of automotive infotainment software.
These actions must be declarative, concise, and tailored to verify the software's functionality accurately.
Test actions should cover marked user interface elements and interactions relevant to the infotainment system.

# Steps
1. Understand UI Element: 
    - Identify the position of the UI element marked by a <color> <marker_type>.
    - Identify the semantic meaning based on its text or icon.
    - Identify any parent elements crucial for semantic or functional meaning.
    - Identify any related elements crucial for semantic or functional meaning (e.g., text corresponding to a checkbox or switch).

2. Reasoning if UI Element is interactive:
    - Determine why the element is interactive or not. Grayed out elements could either mean that they are just disabled or not interactive at all.
    - If it is not interactive set the utterance to "none".

3. Specify Test Action Utterance: 
    - Write a deterministic and declarative action simulating a user interaction with the UI element.
    - Use unique identifiers for the UI element, and mention a parent element if necessary.
    - Do not use the <color> <marker_type> in the test action utterance.
    - Use verbs like tap, click, enter, open, enable, disable, activate, deactivate, select, press, collapse, navigate, cancel, refresh, ...
    - Try to choose the verb that is most applicable for the type of UI Element (e.g., enable for switch, choose for radio buttons, open for submenu etc.)


# Required Output Structure
```
REASONING:
1. [First step in thinking process]
2. [Second step in thinking process]
[Continue with numbered steps as needed]
    
UTTERANCE:
[Describing the test action utterance as a single sentence without using the <color> <marker_type>]
```
\end{lstlisting}

\newpage
\subsubsection{Expected Result Template ``Passed''}    
\label{promt:synt_expected_result_passed}    
\begin{lstlisting}[style=myPromtStyle, basicstyle=\tiny]
As a test engineer for automotive infotainment systems, formulate result evaluations based on the current screen and determine if the test has passed or failed.

# Steps

0. Understand the UI Element:
    - Analyze the current context and active menu of the infotainment system.
    - Identify the position of the UI element marked by a <color> <marker_type>.
    - Determine the semantic meaning based on its text or icon.
    - Identify any parent elements crucial for semantic or functional meaning.
    - Identify any related elements crucial for semantic or functional meaning (e.g., text corresponding to a checkbox or switch).

1. Performed Test Action:
    - Think of a possible test action which was done and can be evaluation with the element marked by the <color> <marker_type>.

2. Reasoning:
    - Conduct reasoning for reaching the result evaluation by examining UI semantics with step-by-step thinking.

3. Determine Evaluation/Expected Result:
    - Provide a short and general description of the expected result of the test action that led to the current screen or highlighted UI element.
    - The expected result must be based solely on the current screen, not on previous or next screens.
    - Include presence, color, positional, semantic, state, and visual information as needed. Do not include the <color> <marker_type>.
    - The expected result can also just check the presence of the marked UI element

4. Incorporate Evaluation:
    - Assess why the expected result is met or not. Conclude with "FAILED" or "PASSED."

# Critical Rules
1. Never reference <color> <marker_type> in test action or expected result
2. Evaluate only current screen state
3. No assumptions about previous/future states
4. Use objective, verifiable statements
5. Document all reasoning steps
6. Provide clear pass/fail criteria
7. Valid evaluations include checking the presence, visibility, position, and properties of the UI element.

# Required Output Structure
```
TEST ACTION:
[Single sentence describing the specific test action performed]
    
REASONING:
1. [First step in thinking process]
2. [Second step in thinking process]
[Continue with numbered steps as needed]
    
EXPECTED RESULT:
[Describing the evaluated result as a single sentence without using the <color> <marker_type>]
    
CONCLUSION:
[PASSED/FAILED]
```
\end{lstlisting}
    
\newpage
\subsubsection{Expected Result Template ``Failed''}
\label{promt:synt_expected_result_failed}
\begin{lstlisting}[style=myPromtStyle, basicstyle=\tiny]
As a test engineer for automotive infotainment systems, formulate result evaluations based on the current screen and determine if the test has passed or failed.

# Steps

1. Understand the UI Element:
    - Analyze the current context and active menu of the infotainment system.
    - Identify the position of the UI element marked by a <color> <marker_type>.
    - Determine the semantic meaning based on its text or icon.
    - Identify any parent elements crucial for semantic or functional meaning.
    - Identify any related elements crucial for semantic or functional meaning (e.g., text corresponding to a checkbox or switch).


2. Determine Evaluation/Expected Result that is wrong for the current screen:
    - Provide a short and general description of the failed expected result of the test action that led to the current screen or highlighted UI element.
    - You should think of an expectation that is wrong or not in the screen.
    - The expected result must be based solely on the current screen, not on previous or next screens.
    - Include absence, different color, wrong positional, semantic, and wrong state information as needed. Do not include the <color> <marker_type>.
    - The expected result can also just check the absence of the marked UI element or if the screen shows the wrong context menue.

3. Reasoning:
    - Conduct reasoning for reaching the result evaluation by examining UI semantics with step-by-step thinking.

4. Incorporate Evaluation:
    - Assess why the expected result is met or not. Conclude with "FAILED" or "PASSED."

# Critical Rules
1. Never reference <color> <marker_type> in test action or expected result
2. The Expectation must be generated for elements within the <color> <marker_type>
3. No assumptions about previous/future states
4. Use objective, verifiable statements
5. Document all reasoning steps
6. Provide clear pass/fail criteria
7. Valid evaluations include checking the presence, visibility, position, and properties of the UI element

# Required Output Structure
```
REASONING:
1. [First step in thinking process]
2. [Second step in thinking process]
[Continue with numbered steps as needed]
    
EXPECTED RESULT:
[Describing the evaluated result as a single sentence without using the <color> <marker_type>]
    
CONCLUSION:
[PASSED/FAILED]
```
\end{lstlisting}

\newpage
\subsection{Examples of Errors for ELAM and AutomotiveUI-Bench-4K: Test Actions}\label{sec:examples_of_visual_errors_ta}

\begin{enumerate}[noitemsep, topsep=0pt]
	\item \begin{minipage}[t]{\linewidth}
		Box too small: \textit{``select kWh/100mi as electric consumption unit"}
		\begin{itemize}
			\item Figure \ref{fig:convertedpic202408091104233021} shows the electric consumption menu of a BMW iX2 in English, featuring a dark background with light text. This menu allows you to change the electric consumption unit. Currently, the unit is set to kWh/100 km.
			\item \ac{elam} is asked to change the unit to kWh/100 mi.
			\item The red box indicates the expected tap area defined in the \textbf{AutomotiveUI-Bench-4K} dataset. The red dot marks the point at which \ac{elam} would tap to perform the test action. Since the red dot is outside the box, the test is counted as failed.
			\item Either the defined area is too small. It should extend to the full length and height of the line containing the radio button for kWh/100 mi, as tapping anywhere on that line activates the button. Therefore, the model did not actually make a mistake. Alternatively, the test action utterance could also be specify to explicitly tap the text or the radio button.
		\end{itemize}
		\begin{center}
			\includegraphics[width=0.85\linewidth]{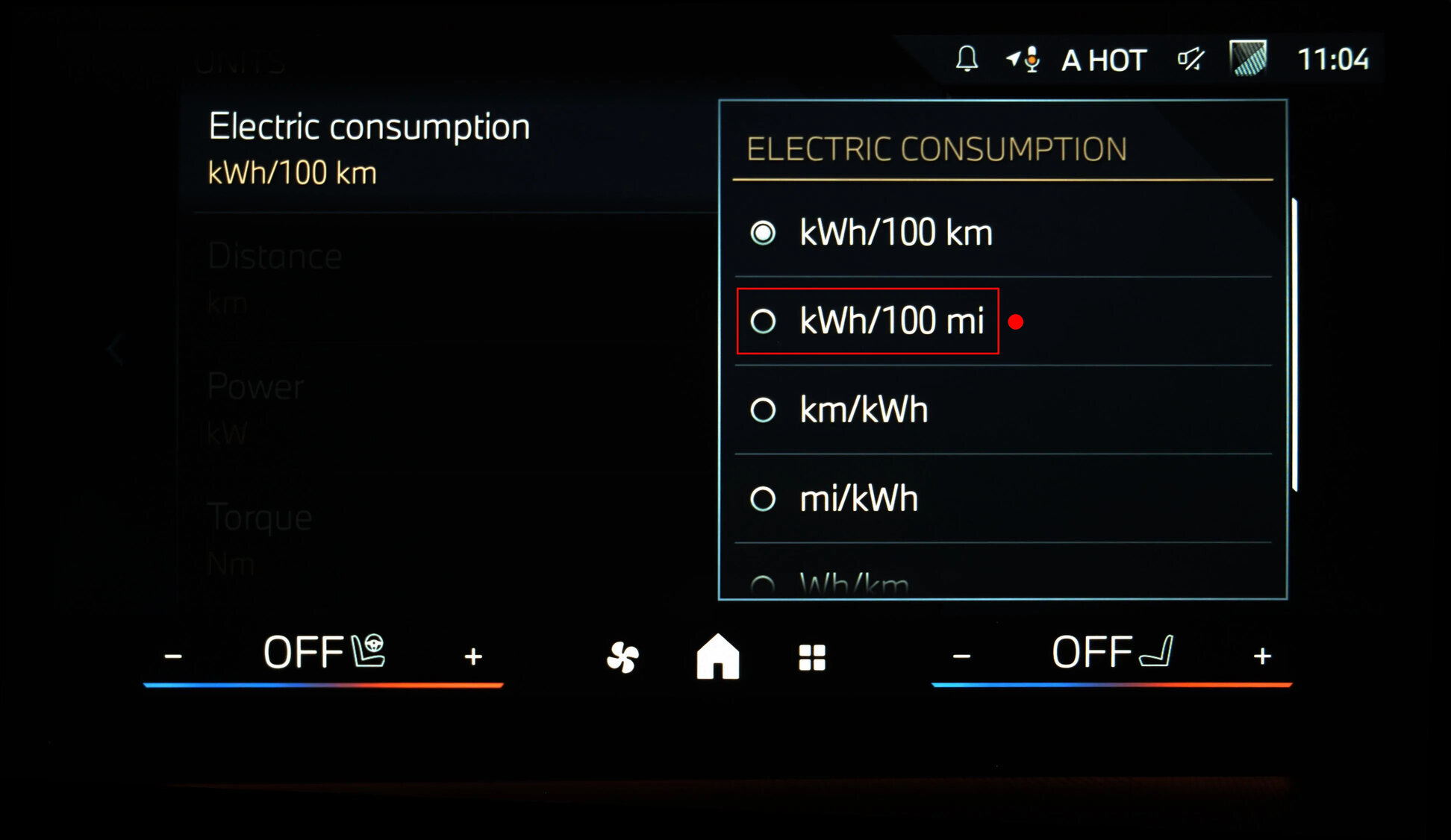}
			\captionof{figure}{Box too small: \textit{``select kWh/100mi as electric consumption unit"}}
			\label{fig:convertedpic202408091104233021}
		\end{center}
	\end{minipage}
	\newpage 
	\item \begin{minipage}[t]{\linewidth}
		Ambiguous or poor description of the test action: \textit{``Select the burger button"}
		\begin{itemize}
			\item Figure \ref{fig:2024-09-051057342802} shows the navigation screen of a Toyota Yaris in German, featuring a dark background with blue buttons and text. A small popup with the text \textit{``Guten Morgen..."} (good morning) is visible in the lower left corner.
			\item \ac{elam}'s task is to select the burger button.
			\item The red box indicating the expected click area marks a button with a hamburger icon in the popup.
			\item The issue lies in the presence of two additional \textit{``burger buttons"} that are currently visible on the screen. There is a button with a burger icon and the text \textit{``Lebensmittel"} (groceries) at the bottom of the screen and there is a button featuring an icon showing three horizontal like, which is generally known as \textit{``burger button"}, at the upper right corner of the screen.
		\end{itemize}
		\begin{center}
			\includegraphics[width=0.85\linewidth]{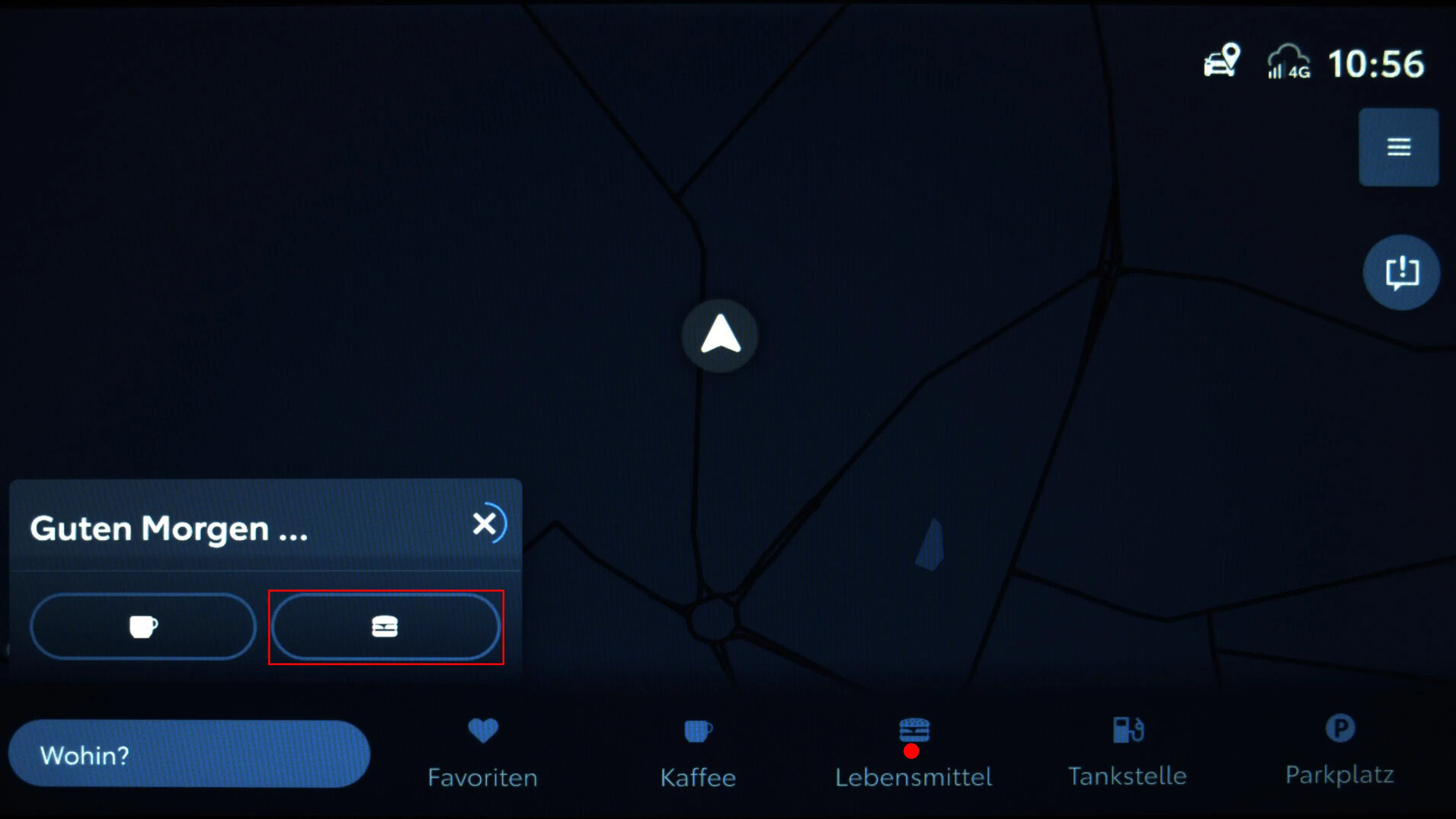}
			\captionof{figure}{Ambiguous or poor description of the test action: \textit{``Select the burger button"}}
			\label{fig:2024-09-051057342802}
		\end{center}
	\end{minipage}
	
	\item \begin{minipage}[t]{\linewidth}
		Special icons from the vehicle domain: \textit{``turn steering wheel heating on"}
		\begin{itemize}
			\item Figure \ref{fig:convertedpic202409171422305696} shows the navigation screen of a Ford Mustang Mach-E with a white background. The area of interest in this test is the lower quarter of the screen featuring buttons related to climate control.
			\item \ac{elam} is asked to turn on the steering wheel heating.
			\item The steering wheel heating icon is marked with a red box in the lower left corner of the screen. The small gray line below the icon indicates that the heating is currently turned off.
			\item \ac{elam} tapped the multi-zone climate control icon instead, which features a windshield heating symbol. The steering wheel heating icon is a car-specific icon, \ac{elam} would benefit from more training data with car-specific icons.
		\end{itemize}
		\begin{center}
			\includegraphics[width=0.5\linewidth]{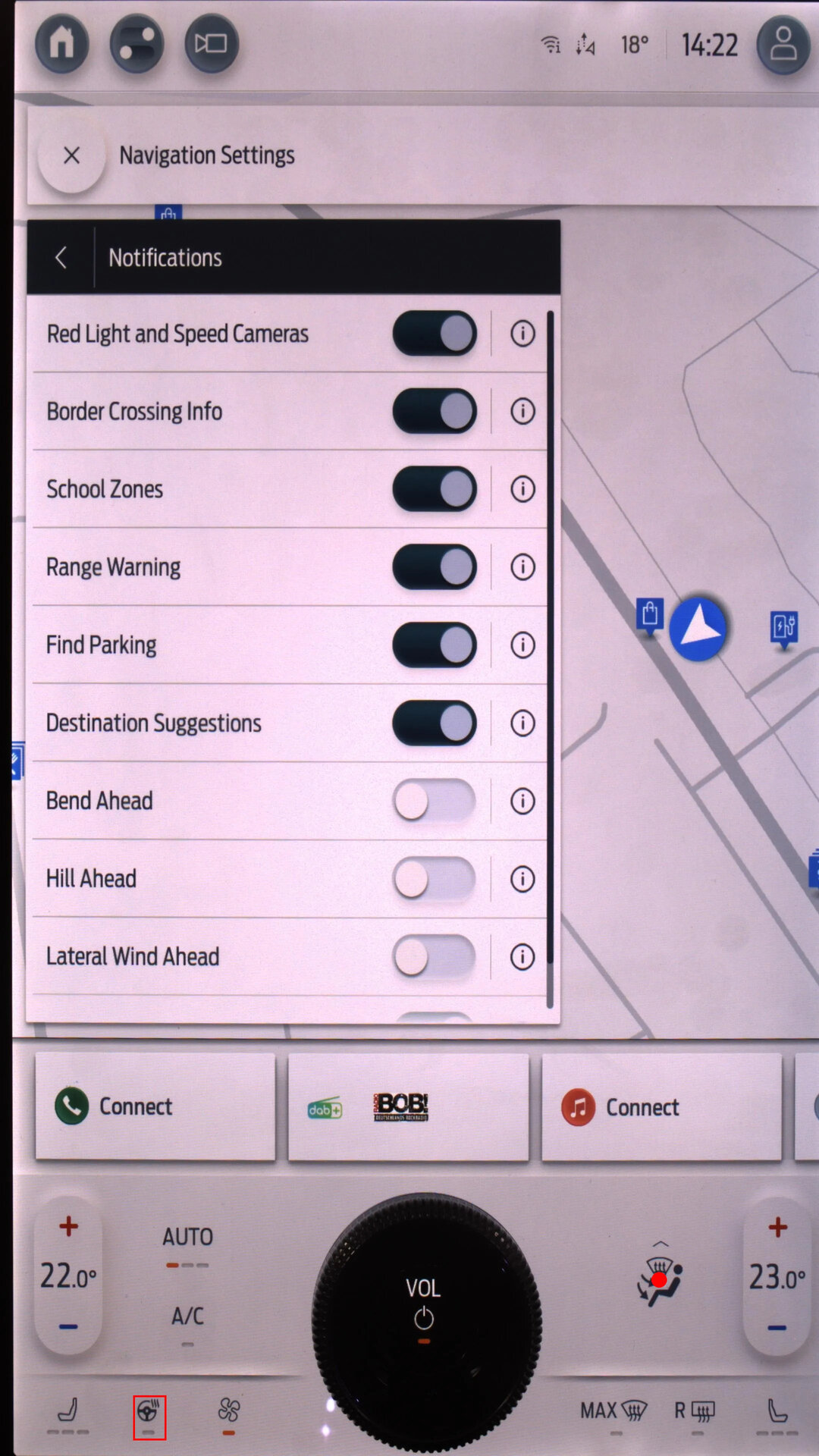}
			\captionof{figure}{Special icons from the vehicle domain: \textit{``turn steering wheel heating on"}}
			\label{fig:convertedpic202409171422305696}
		\end{center}
	\end{minipage}
	
	\item \begin{minipage}[t]{\linewidth}
		Clear description, but very similar elements confused the model: \textit{``deactivate the reminder signal for mobile phone"}
		\begin{itemize}
			\item Figure \ref{fig:convertedpic202409061112523142} shows the settings menu of an Audi e-tron GT, featuring a black background with white text.
			\item The test action is to deactivate the reminder signal for mobile phone.
			\item The red box surrounds the \textit{``Off"} button below the text \textit{``Reminder signal for mobile phone"}. The value is currently set to \textit{``Spoken"}, as indicated by the white line at the bottom of the button.
			\item However, \ac{elam} tapped the toggle button next to \textit{``Mobile phone notifications: Reminder/charge level"}. The test action clearly states that \textit{``Reminder signal for mobile phone"} should be deactivated and not \textit{``Mobile phone notifications: Reminder/charge level"}. The issue lies in the text's close semantic resemblance, which led to confusion in the model.
		\end{itemize}
		\begin{center}
			\includegraphics[width=0.85\linewidth]{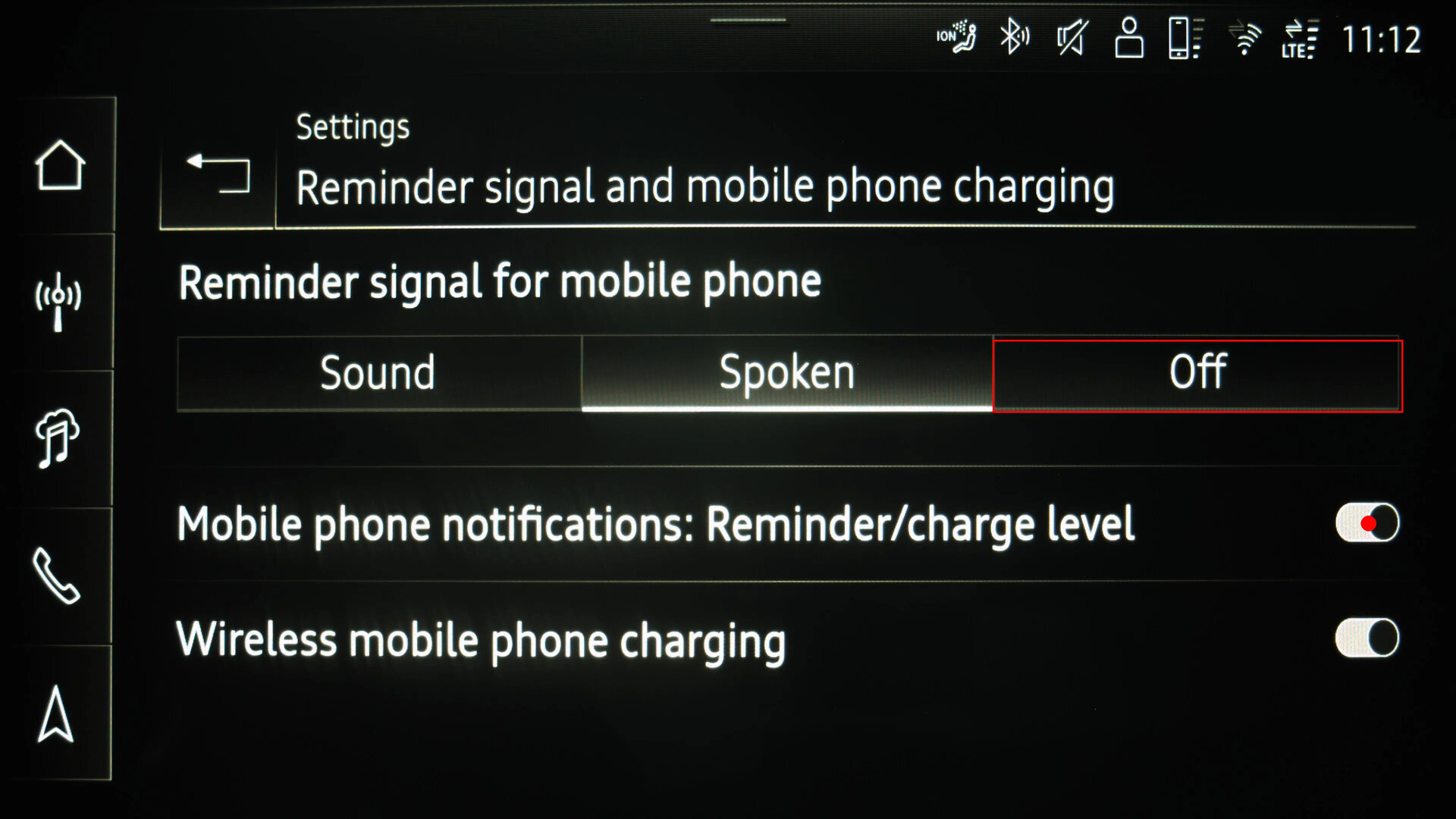}
			\captionof{figure}{Clear description, but very similar elements confused the model: \textit{``deactivate the reminder signal for mobile phone"}}
			\label{fig:convertedpic202409061112523142}
		\end{center}
	\end{minipage}
	
	\item \begin{minipage}[t]{\linewidth}
		Connection between elements could not be established: \textit{``add phone to Favourites"}
		\begin{itemize}
			\item Figure \ref{fig:convertedpic202409231407459594} shows the apps menu of a Maserati Grecale with a black background and white icons and text. Eight apps are currently visible. Each app features a star icon to add it to favourites.
			\item The task is to add phone to the favourites.
			\item The star icon next to the phone button is marked by the red box, which indicates that it is the expected aria to click.
			\item The red dot indicates that \ac{elam} intends to tap directly on the phone button instead of the star icon which would add phone to favourites. \ac{elam} was not capable of making a connection between the phone button and the star icon.
		\end{itemize}
		\begin{center}
			\includegraphics[width=0.85\linewidth]{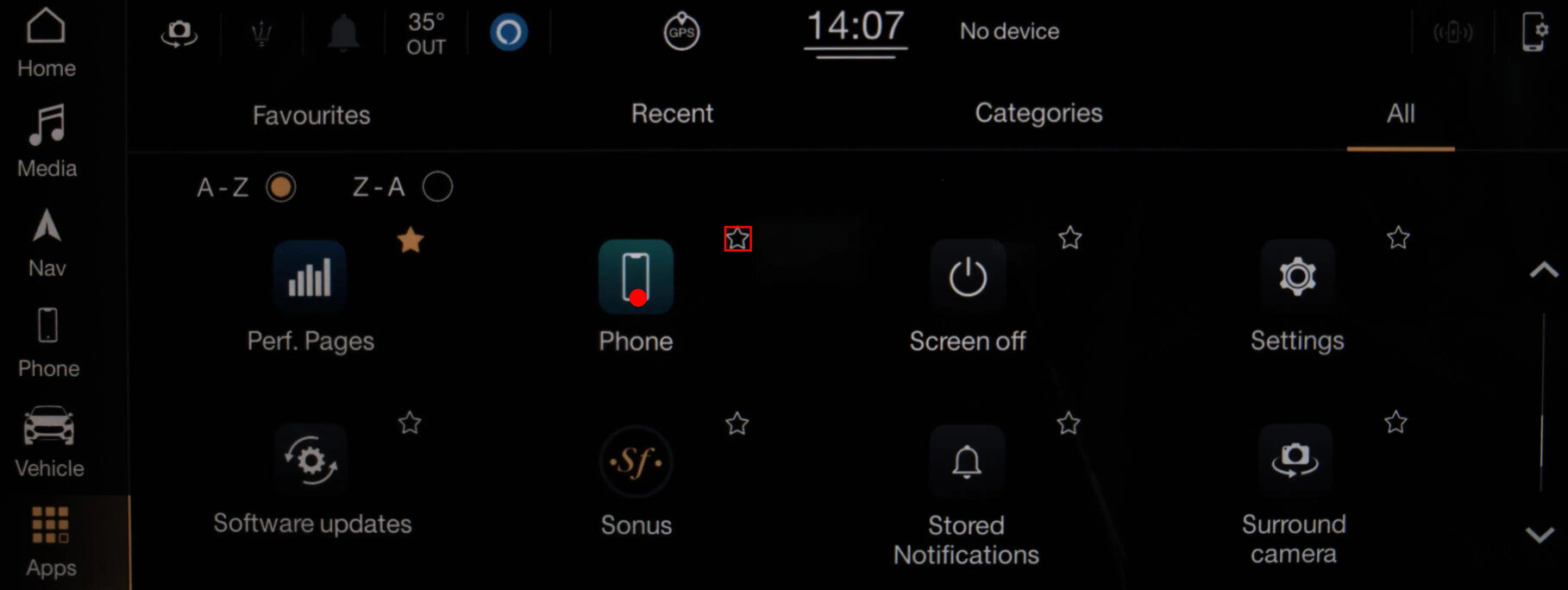}
			\captionof{figure}{Connection between elements could not be established: \textit{``add phone to Favourites"}}
			\label{fig:convertedpic202409231407459594}
		\end{center}
	\end{minipage}
	
	\item \begin{minipage}[t]{\linewidth}
		Driver/passenger, left/right, up/down distinctions: \textit{``increase the right temperature setting with the plus button"}
		\begin{itemize}
			\item Figure \ref{fig:2024-09-091107413591} shows the climate menu of an Opel Astra in German. The background color is a very dark red and text and icons are white. Some icons/texts are highlighted in orange.
			\item \ac{elam} is asked to increase the right temperature setting with the plus button.
			\item The expected click area is marked with a red box on the right side of the screen (plus button).
			\item \ac{elam} selected the left plus button instead of the right plus button. The underlying cause of the issue can be traced back to the training data. In instances where sufficient data is not available for left versus right comparisons, the model is unable to differentiate between these two options.
		\end{itemize}
		\begin{center}
			\includegraphics[width=0.85\linewidth]{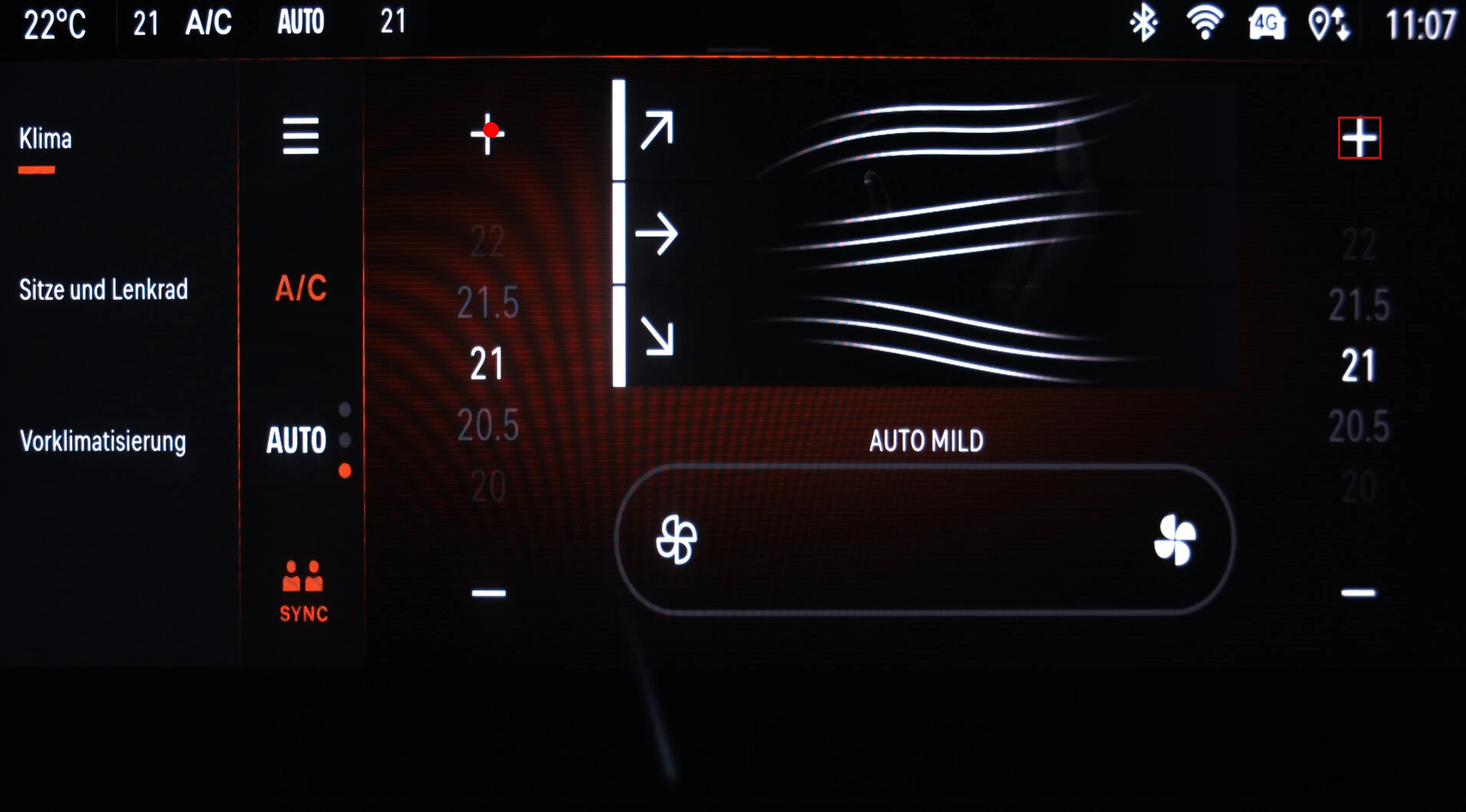}
			\captionof{figure}{Driver/passenger, left/right, up/down distinctions: \textit{``increase the right temperature setting with the plus button"}}
			\label{fig:2024-09-091107413591}
		\end{center}
	\end{minipage}
	
	\item \begin{minipage}[t]{\linewidth}
		Elements must be counted: \textit{``Activate the first weekly item from the charging list"}
		\begin{itemize}
			\item Figure \ref{fig:convertedpic202409091302133680} shows the charging menu of a Mini Cooper in German. The round screen is unusual for a car's infotainment system. The menu allows the user to set several timers for charging.
			\item The task is to activate the first weekly timer.
			\item The toggle button of the first weekly item in the list is marked by a red box, indicating the expected tap area.
			\item \ac{elam} selected the second weekly item. This suggests that there was insufficient training data containing counting examples to teach the model how to count.
		\end{itemize}
		\begin{center}
			\includegraphics[width=0.5\linewidth]{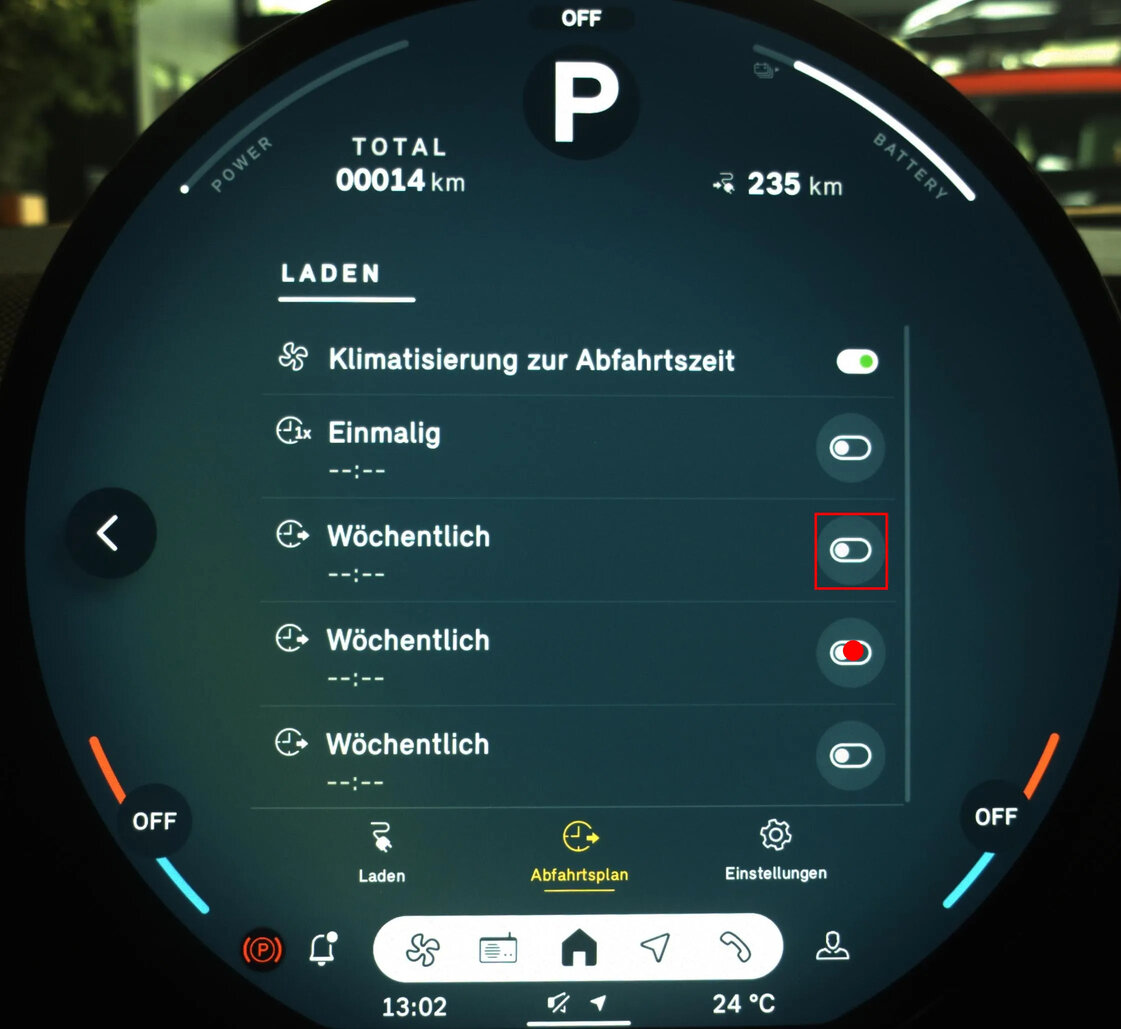}
			\captionof{figure}{Elements must be counted: \textit{``Activate the first weekly item from the charging list"}}
			\label{fig:convertedpic202409091302133680}
		\end{center}
	\end{minipage}
	
	\item \begin{minipage}[t]{\linewidth}
		Miscellaneous: \textit{``Go to main menu"}
		\begin{itemize}
			\item Figure \ref{fig:convertedpic202409121047118181} shows the camera settings menu of a Kia. This infotainment system has a light background with dark text and bluish highlights.
			\item \ac{elam} is asked to go to the main menu.
			\item The red box, situated around the house icon that is visible at the top of the screen, serves as an indicator of the designated tap area.
			\item \ac{elam} taps on the context menu burger button instead of the house icon, despite the fact that a house icon is commonly used as an icon for the main menu, as is also the case outside of the car domain. The rationale behind the models decision to tap there remains unclear.
		\end{itemize}
		\begin{center}
			\includegraphics[width=0.85\linewidth]{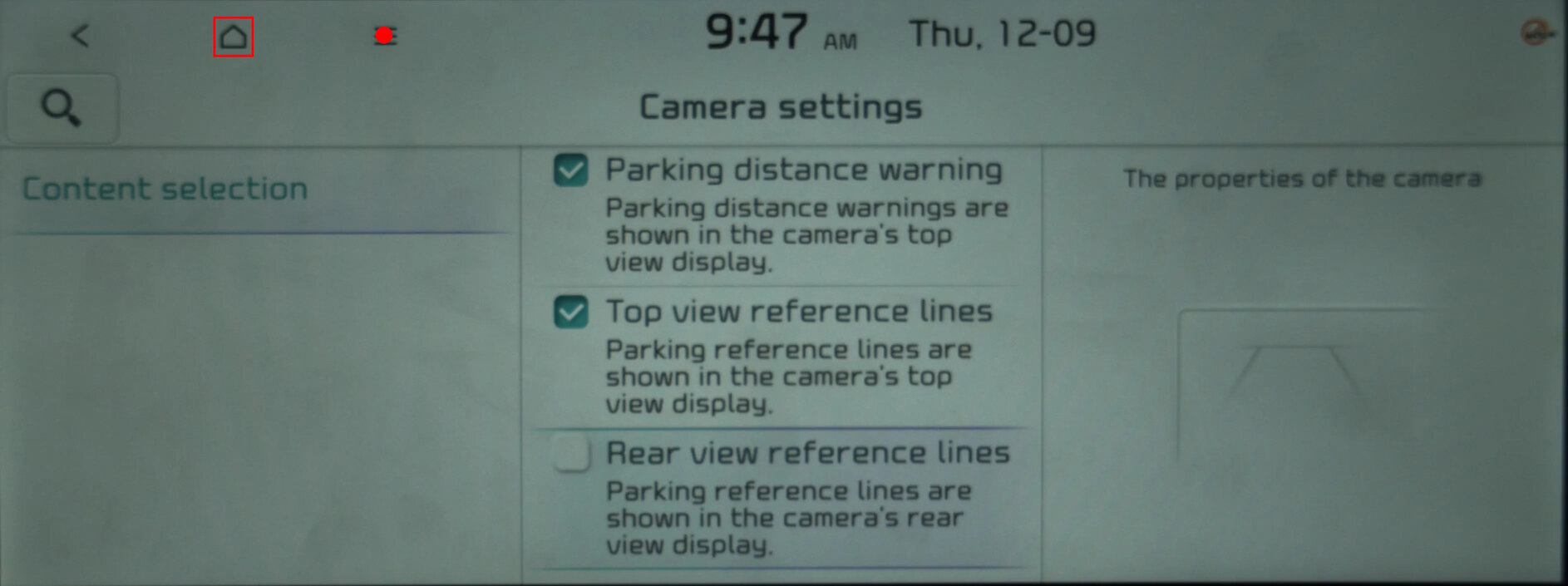}
			\captionof{figure}{Miscellaneous: \textit{``Go to main menu"}}
			\label{fig:convertedpic202409121047118181}
		\end{center}
	\end{minipage}
\end{enumerate}

\subsection{Examples of Errors for ELAM and AutomotiveUI-Bench-4K: Expected Results}\label{sec:examples_of_visual_errors_er}

\begin{enumerate}[noitemsep, topsep=0pt]
	\item \begin{minipage}[t]{\linewidth}
		Box too small: \textit{``Audio quality is set to low"}
		\begin{itemize}
			\item Figure \ref{fig:convertedpic202409101018440791} shows the audio settings menu of a Cupra Leon in German, featuring a dark background with light text. Selected texts are highlighted in orange.
			\item \ac{elam} is asked to check if audio quality is set to low. The expected answer is \textbf{failed}. \ac{elam} answers correctly with \textbf{failed}.
			\item The red box indicates the region defined within the image in the AutomotiveUI-Bench-4K dataset where the result is visible. The red dot marks the point at which \ac{elam} focused to determine the result. Since the red dot is outside the box, the test is counted as failed, even though the result was correct.
			\item In this case, the defined area is too small. It should extend to the full length and height of the line containing the selected option \textit{"Hoch"} (high).
		\end{itemize}
		\begin{center}
			\includegraphics[width=0.85\linewidth]{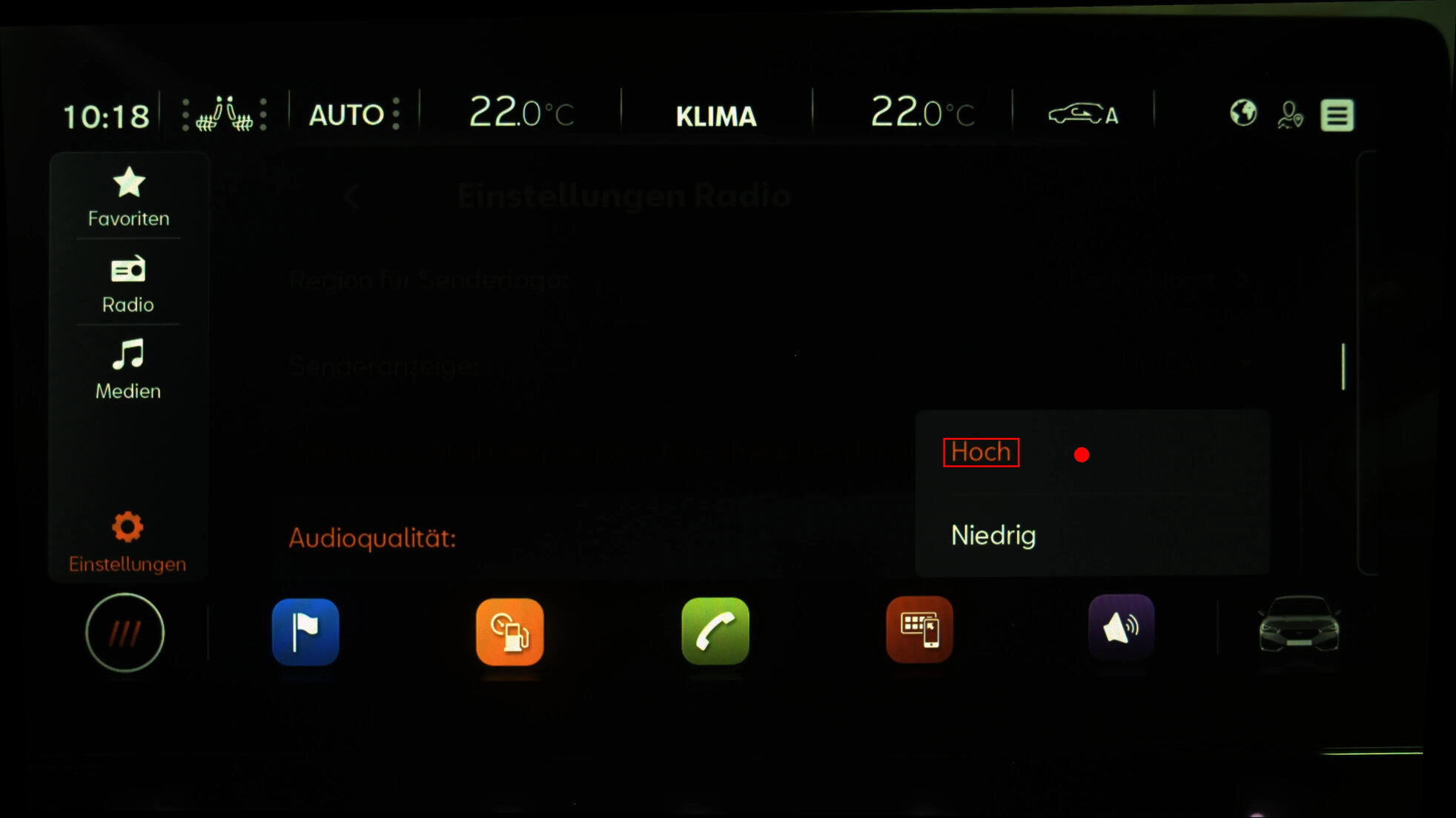}
			\captionof{figure}{Box too small: \textit{``Audio quality is set to low"}} - expected result: \textbf{failed}, \ac{elam}'s prediction: \textbf{failed}
			\label{fig:convertedpic202409101018440791}
		\end{center}
	\end{minipage}
	\newpage 
	\item \begin{minipage}[t]{\linewidth}
		Multiple areas display the expected result: \textit{``Sound settings are displayed"}
		\begin{itemize}
			\item Figure \ref{fig:convertedpic202408091051069353} shows the sound settings menu of a BMW iX2 in English, featuring a dark background with white icons and text. Selected items are highlighted in light yellow.
			\item The utterance to verify is: Sound settings are displayed. \ac{elam} answers correctly with \textbf{passed}.
			\item The red box indicating the area of focus of \ac{elam} marks the sound settings tab on the left, but the region surrounded by the red box is the heading of the sound menu at the top of the page.
			\item Both regions show the expected result, meaning there was only a 50\% chance that \ac{elam} would point to the correct region. However, drawing a bigger box around both regions is problematic because it could lead to false positives. Either the dataset needs the ability to store multiple regions for one utterance, or the utterance needs to be more precise, for example: \textit{"The sound settings are shown under the SOUND label located on the left side of the top status bar."}
		\end{itemize}
		\begin{center}
			\includegraphics[width=0.85\linewidth]{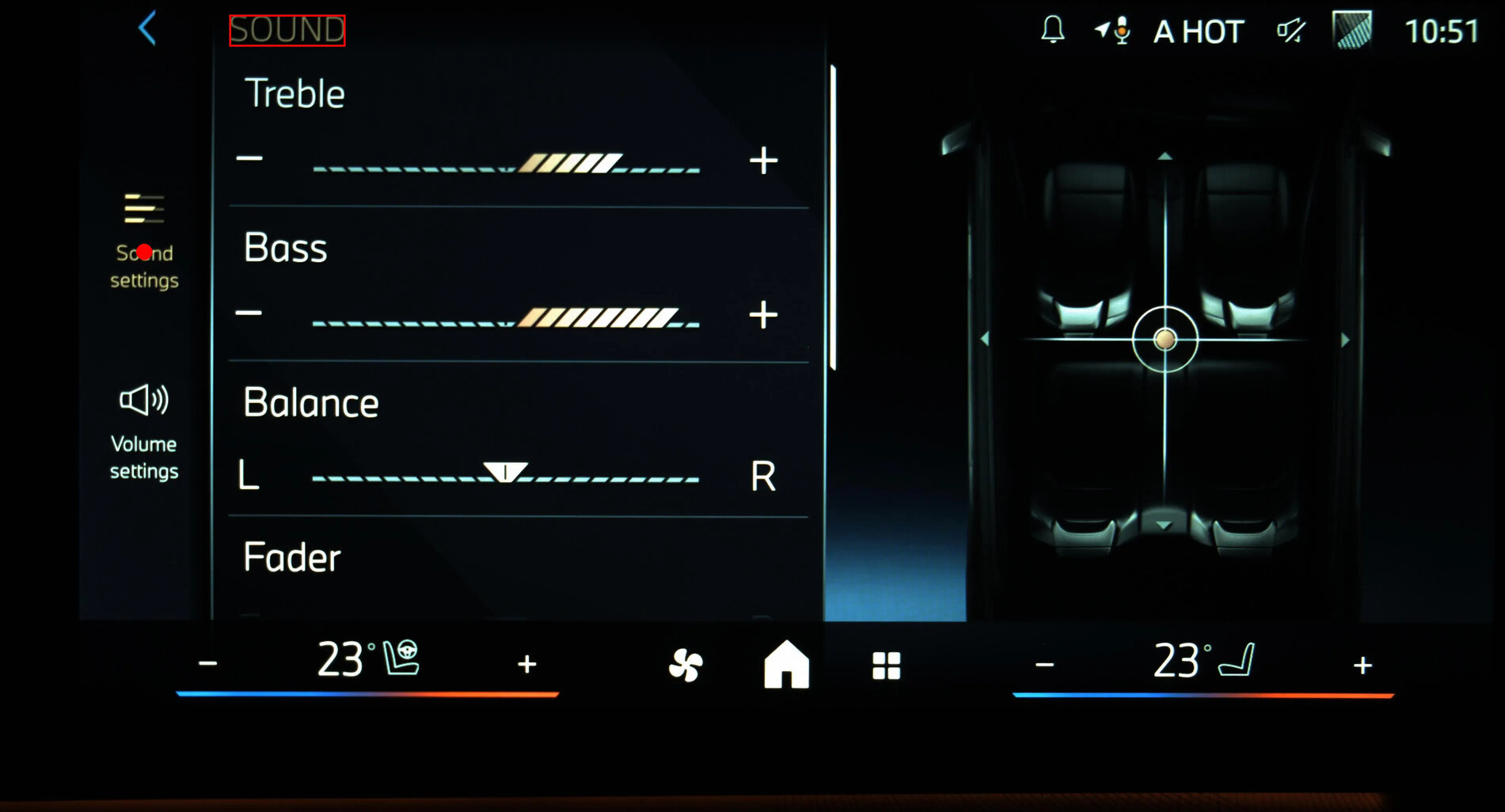}
			\captionof{figure}{Multiple areas display the expected result: \textit{``Sound settings are displayed"}} - expected result: \textbf{passed}, \ac{elam}'s prediction: \textbf{passed}
			\label{fig:convertedpic202408091051069353}
		\end{center}
	\end{minipage}
	
	\item \begin{minipage}[t]{\linewidth}
		Box not around option, when asked if options are shown: \textit{``Driver assistance options are displayed"}
		\begin{itemize}
			\item Figure \ref{fig:convertedpic202409121046190231} shows the driver assistance menu of a Kia with a white background. The forward safety settings are currently selected.
			\item \ac{elam} needs to check if the driver assistance options are displayed.
			\item The menu heading \textit{``Driver assistance"} is marked with a red box. However, \ac{elam}'s area of interest was the section below the heading where the menu options are displayed.
			\item Since emphasis was placed on the fact that options are displayed, not on the menu heading being \textit{``Driver assistance"}, the box was drawn around the wrong region and the data should be corrected in \textbf{AutomotiveUI-Bench-4K}.
		\end{itemize}
		\begin{center}
			\includegraphics[width=0.85\linewidth]{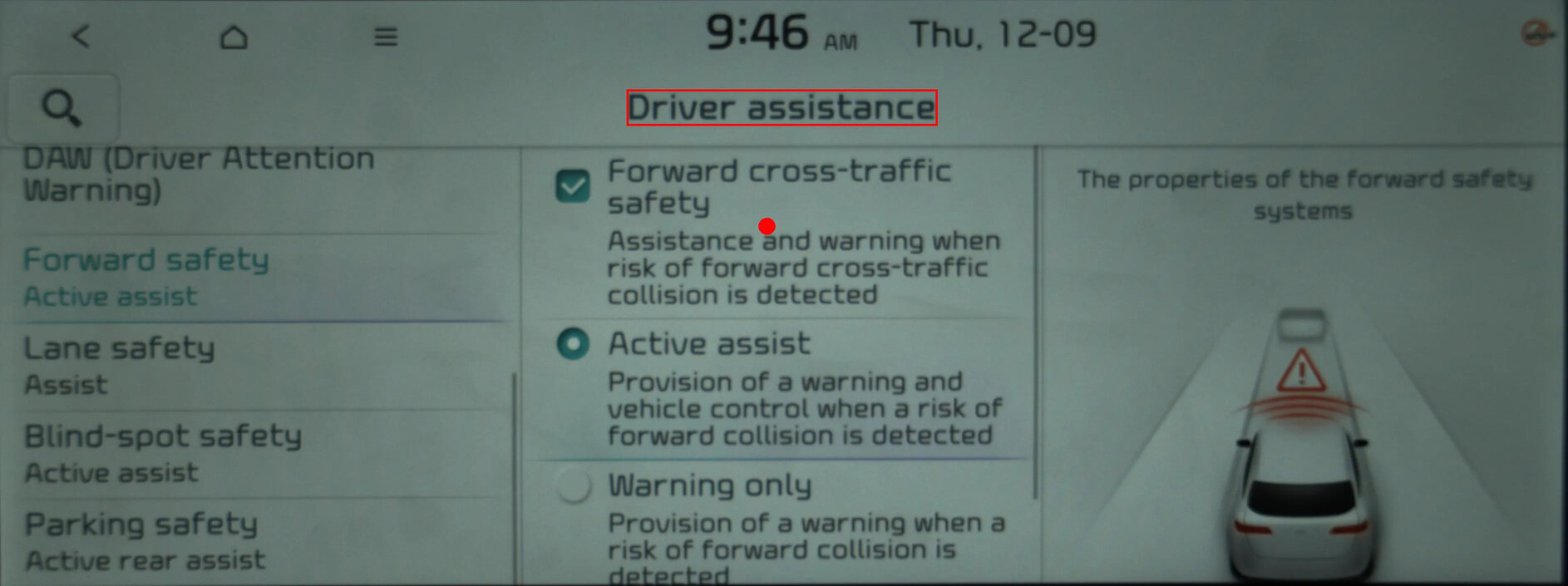}
			\captionof{figure}{Box not around option, when asked if options are shown: \textit{``Driver assistance options are displayed"}} - expected result: \textbf{passed}, \ac{elam}'s prediction: \textbf{passed}
			\label{fig:convertedpic202409121046190231}
		\end{center}
	\end{minipage}
	
	\item \begin{minipage}[t]{\linewidth}
		Special icons from the vehicle domain: \textit{``The medium sensitivity of the distance control was selected"}
		\begin{itemize}
			\item Figure \ref{fig:convertedpic202408091036435144} shows the distance control menu of a BMW iX2 in German with a black background and light text.
			\item The expected results utterance is: The medium sensitivity of the distance control was selected. \ac{elam} answer is \textbf{failed} instead of \textbf{passed}.
			\item The red box surrounds the medium sensitivity icon. \ac{elam} selected the toggle button next to \textit{``Wechsel zu Geschwindigkeitsregelung''} (switch to speed control).
			\item The the medium sensitivity icon with the two cars and two lines in between is a very automotive-specific icon. Because \ac{elam} did not recognize the icon, it pointed to the switch to speed control setting, probably because the explanation below also contains the word distance control (\textit{``Abstandsregelung''}).
		\end{itemize}
		\begin{center}
			\includegraphics[width=0.85\linewidth]{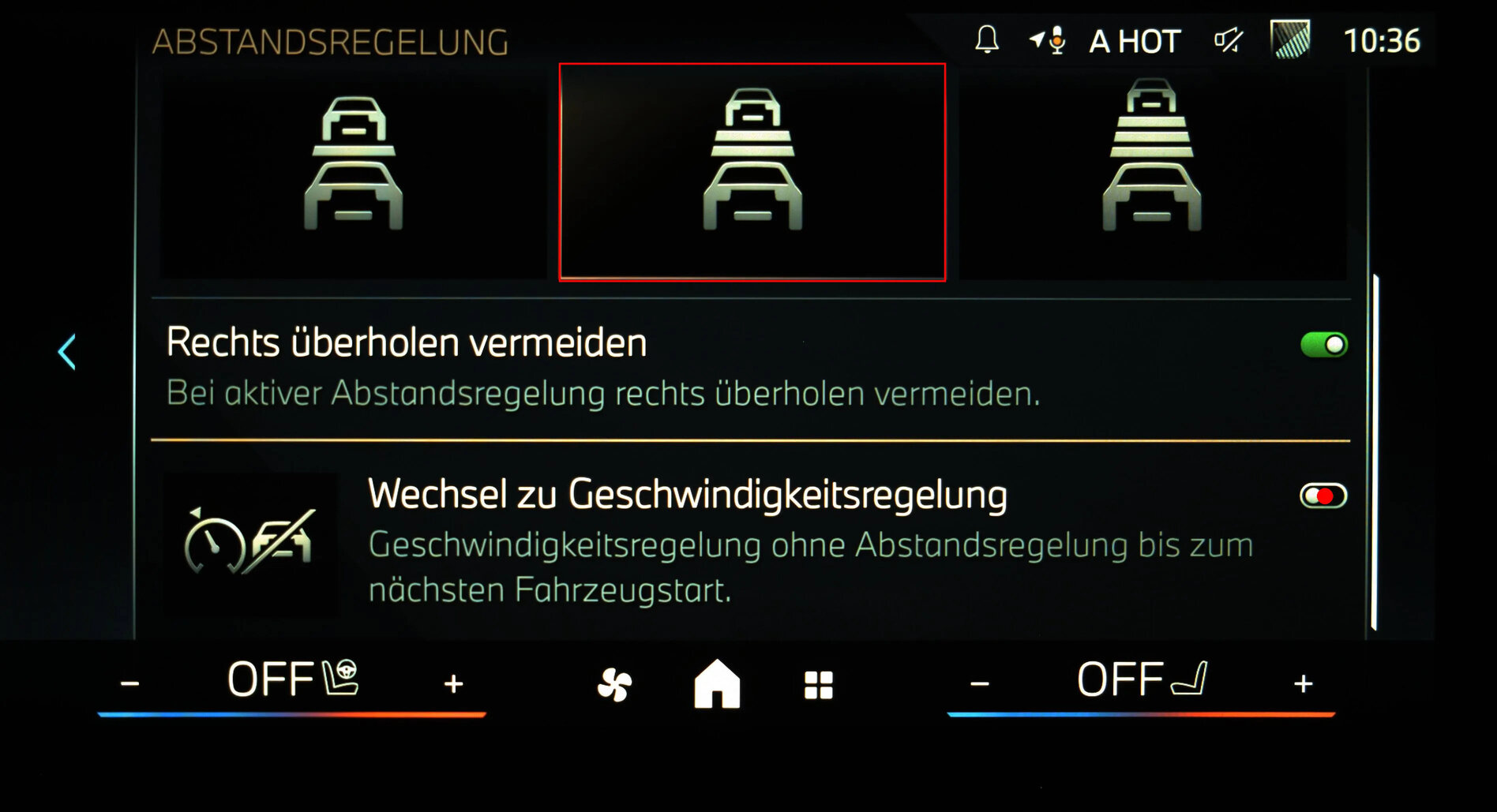}
			\captionof{figure}{Special icons from the vehicle domain: \textit{``The medium sensitivity of the distance control was selected"}} - expected result: \textbf{passed}, \ac{elam}'s prediction: \textbf{failed}
			\label{fig:convertedpic202408091036435144}
		\end{center}
	\end{minipage}
	
	\item \begin{minipage}[t]{\linewidth}
		Clear description, but very similar elements confused the model: \textit{``Noise reduction is disabled"}
		\begin{itemize}
			\item Figure \ref{fig:convertedpic202409121108512250} shows the audio settings menu of a Kia in German, featuring a light background with gray text.
			\item The utterance to verify is: Noise cancellation is disabled. The expected result is \textbf{passed} but \ac{elam}'s answer was \textbf{failed}.
			\item The red box surrounds the first radio button item in the list, that belongs to the noise reduction settings, labeled \textit{``Originalklang"} (original sound). \ac{elam} suggests that the second item labeled \textit{``Leichte Rauschunterdr\"{u}ckung"} (light noise reduction) is the item of interest because it corresponds more to \textit{noise reduction} semantically than \textit{``original sound"}.
			\item However, the explanation provided below the radio button indicates that when the first item is selected, no noise reduction is applied at all.
		\end{itemize}
		\begin{center}
			\includegraphics[width=0.85\linewidth]{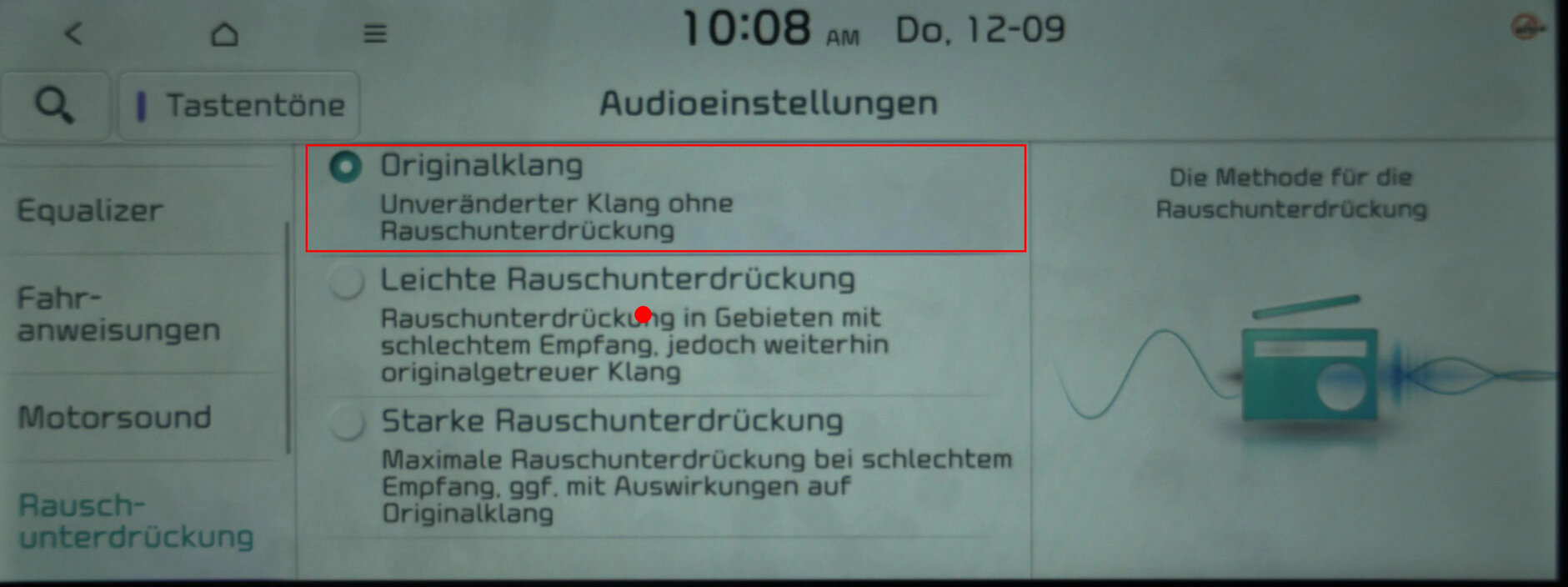}
			\captionof{figure}{Clear description, but very similar elements confused the model: \textit{``Noise reduction is disabled"}} - expected result: \textbf{passed}, \ac{elam}'s prediction: \textbf{failed}
			\label{fig:convertedpic202409121108512250}
		\end{center}
	\end{minipage}

	\item \begin{minipage}[t]{\linewidth}
		Miscellaneous: \textit{``E-Call settings are shown"}
		\begin{itemize}
			\item Figure \ref{fig:2024-06-041127504700} shows the Wi-Fi menu of a VW ID.4. This infotainment system has a black background with white text.
			\item \ac{elam} is asked to check if emergency call (E-Call) settings are currently displayed. \ac{elam} answers correctly with \textbf{failed}.
			\item The red box is situated around the entire Wi-Fi menu.
			\item Even though \ac{elam} answered the question correctly, its area of focus was not within the Wi-Fi menu but on the left side, on the A/C off indicator.
		\end{itemize}
		\begin{center}
			\includegraphics[width=0.85\linewidth]{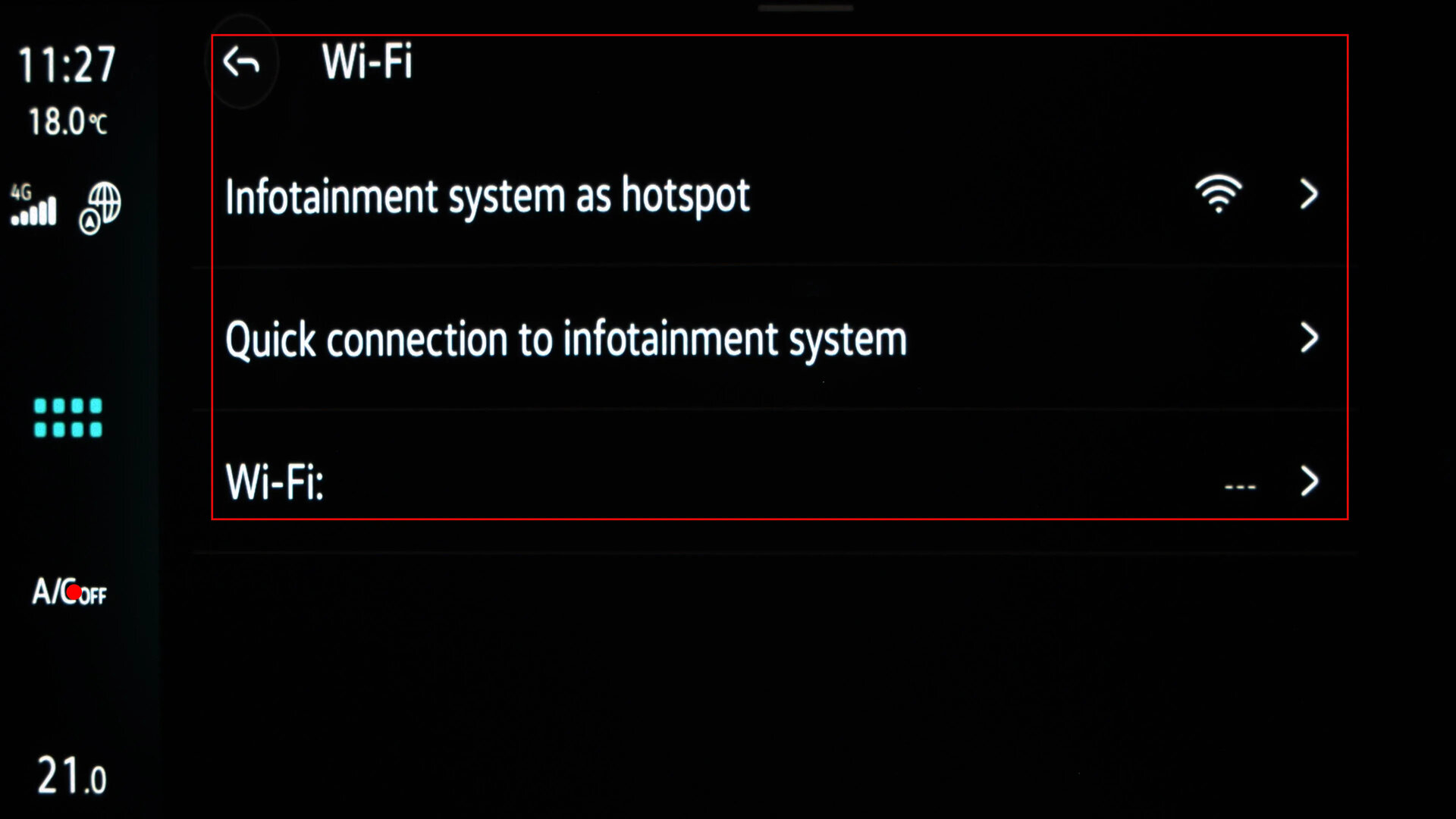}
			\captionof{figure}{Miscellaneous: \textit{``E-Call settings are shown"}} - expected result: \textbf{failed}, \ac{elam}'s prediction: \textbf{failed}
			\label{fig:2024-06-041127504700}
		\end{center}
	\end{minipage}
\end{enumerate}
\end{document}